\documentclass{article}
\usepackage[preprint]{neurips_2024}

\usepackage[utf8]{inputenc}
\usepackage[T1]{fontenc}   
\usepackage{amsfonts}      
\usepackage{microtype}
\usepackage{graphicx}
\usepackage{subcaption}
\usepackage{booktabs}
\usepackage{multirow}
\usepackage{url}
\usepackage{makecell, tabularx, xcolor}
\usepackage{colortbl}
\usepackage{csquotes}
\usepackage{mathtools}
\usepackage{enumitem}

\usepackage{bbding}
\usepackage{pifont}

\newcommand{\cmark}{\ding{51}}
\newcommand{\xmark}{\ding{55}}

\usepackage{graphicx,amsmath,amssymb,amsthm,mathtools,bbm} 
\usepackage{hyperref}
\usepackage[capitalize,noabbrev]{cleveref}

\setlist[itemize]{leftmargin=*}
\setlist[enumerate]{leftmargin=*}

\newcommand{\E}{\mathbbm{E}}
\newcommand{\R}{\mathbb{R}}
\renewcommand{\div}{\operatorname{div}}
\newcommand{\tr}{\operatorname{Tr}}

\makeatletter
\DeclareRobustCommand{\cev}[1]{%
  {\mathpalette\do@cev{#1}}%
}
\newcommand{\do@cev}[2]{%
  \vbox{\offinterlineskip
    \sbox\z@{$\m@th#1 x$}%
    \ialign{##\cr
      \hidewidth\reflectbox{$\m@th#1\vec{}\mkern4mu$}\hidewidth\cr
      \noalign{\kern-\ht\z@}
      $\m@th#1#2$\cr
    }%
  }%
}
\makeatother

\theoremstyle{plain}
\newtheorem{theorem}{Theorem}[section]
\newtheorem{proposition}[theorem]{Proposition}

\theoremstyle{definition}

\theoremstyle{remark}
\newtheorem{remark}[theorem]{Remark}

\hypersetup{
    colorlinks,
    linkcolor={red!50!black},
    citecolor={blue!30!black},
    urlcolor={blue!80!black}
}

\title{Dynamical Measure Transport \\and Neural PDE Solvers for Sampling}

\newcommand*\samethanks[1][\value{footnote}]{\footnotemark[#1]}

\author{Jingtong Sun\thanks{Equal contribution.},\, Julius Berner\samethanks[1] \\
California Institute of Technology\\
\And
Lorenz Richter\samethanks[1]\\
Zuse Institute Berlin \\
dida Datenschmiede GmbH\\
\And
Marius Zeinhofer \\
Simula Research Laboratory \\
University Hospital Freiburg \\
\And
Johannes Müller \\
RWTH Aachen \\
\And
Kamyar Azizzadenesheli\\
NVIDIA\\
\And
Anima Anandkumar\\
California Institute of Technology\\
}

\begin{document}
\maketitle
\begin{abstract}
    The task of sampling from a probability density can be approached as transporting a tractable density function to the target, known as dynamical measure transport. In this work, we tackle it through a principled unified framework using deterministic or stochastic evolutions described by partial differential equations (PDEs). This framework incorporates prior trajectory-based sampling methods, such as diffusion models or Schrödinger bridges, without relying on the concept of time-reversals. Moreover, it allows us to propose novel numerical methods for solving the transport task and thus sampling from complicated targets without the need for the normalization constant or data samples. We employ physics-informed neural networks (PINNs) to approximate the respective PDE solutions, implying both conceptional and computational advantages. In particular, PINNs allow for simulation- and discretization-free optimization and can be trained very efficiently, leading to significantly better mode coverage in the sampling task compared to alternative methods. Moreover, they can readily be fine-tuned with Gauss-Newton methods to achieve high accuracy in sampling. 
\end{abstract}

\section{Introduction}

We consider the problem of sampling from a target probability density
\begin{equation}
    p_{\mathrm{target}} = \frac{\rho_{\mathrm{target}}}{Z}, \qquad Z \coloneqq \int \rho_{\mathrm{target}}(x) \, \mathrm{d}x,
\end{equation}
for which only an unnormalized function $\rho_{\mathrm{target}}\colon \R^d \to (0,\infty)$ can be evaluated, but the normalizing constant $Z$ is typically intractable. This challenging task has wide applications, for instance, in Bayesian statistics \cite{turkman2019computational}, computational physics \cite{stoltz2010free}, quantum chemistry \cite{noe2019boltzmann,kanwar2020equivariant} and other scientific disciplines \cite{glasserman2004monte,mode2011applications}. Various particle-based methods, such as importance sampling, Markov chain Monte Carlo (MCMC), Sequential Monte Carlo, etc., have been designed in the last decades to approach this task \cite{liu2001monte,doucet2001sequential,martino2018independent}. However, they often suffer from slow convergence, in particular for high-dimensional, multimodal distributions. In order to address this issue and improve sampling performance, two paradigms have been introduced:
\begin{enumerate}[itemsep=0.5pt]
    \item Enhancing the sampling problem with a learning task, where usually some function is learned in order to improve sampling quality (e.g., in a variational inference setting).
    \item Formulating the sampling problem as a dynamical measure transport from a tractable initial density function to the complicated target.
\end{enumerate}
In this work, we aim to advance both paradigms. In particular, we rely on the underlying principled framework of partial differential equations (PDEs) as a unified framework for deriving both existing and new sampling algorithms.

To be more precise, we consider the task of identifying evolutions from an initial distribution to the target on a finite-time horizon. There are two broad approaches to tackle this, viz., particle and density-based approaches. Particle-based approaches sample so-called particles from the initial distribution and evolve them using differential equations, either with (deterministic) ordinary differential equations (ODEs) or stochastic differential equations (SDEs). In contrast, for density-based approaches,   the evolutions of the densities (of the particles) can be described by associated PDEs, viz., the continuity or the Fokker-Planck equation, respectively. In particular, the PDEs couple the drift of the ODE or SDE and the density, giving us the choice to add additional constraints (leading to unique optimal values) or to learn both simultaneously (leading to non-unique solutions). While unique solutions can exhibit beneficial properties (such as drifts with small magnitude), the existence of multiple optimal solutions can be more suited for gradient-based optimization methods.

For the task of numerically approximating the high-dimensional PDEs at hand, we can leverage different deep-learning methods. We show that we can recover multiple previous methods when considering losses based on backward stochastic differential equations (BSDEs). This highlights the foundational role of the PDE framework \cite{han2017deep,nusken2021interpolating}.
Employing the framework of physics-informed neural networks (PINNs) \cite{raissi2017physics}, we derive novel variational formulations with both unique and non-unique solutions.
More importantly, the PINN losses only require evaluating the PDE residual on random points in the spatio-temporal domain. In contrast, previous works based on dynamical measure transport rely on discretized trajectories of the dynamics for training. We numerically evaluate our PINN-based approaches on challenging high-dimensional examples and show better performance. In particular, we can improve mode coverage in multimodal settings compared to simulation-based approaches.

Our contributions can be summarized as follows:
\begin{itemize}[itemsep=0.5pt]
    \item We provide a unifying PDE perspective on generative modeling and sampling via dynamical measure transport. 
    \item We derive suitable objectives to numerically solve these PDEs using deep learning. This recovers known methods as special cases and provides a range of novel objectives with beneficial numerical properties.  
    \item We propose further improvements based on efficient parametrizations, sampling schemes, and optimization routines. This leads to state-of-the-art performance on a series of benchmarks.
\end{itemize}

\subsection{Related work}
\label{sec:related}

There are numerous Monte Carlo-based methods for sampling from unnormalized densities, including \emph{Markov chain Monte Carlo} (MCMC) \citep{kass1998markov}, \emph{Annealed Importance Sampling} (AIS) \citep{neal2001annealed}, and \emph{Sequential Monte Carlo} (SMC) \citep{del2006sequential,doucet2009tutorial}. However, these methods typically only guarantee \emph{asymptotic} convergence to the target density, with potentially slow convergence rates in practical scenarios \citep{robert1999monte}. Variational methods, such as \emph{mean-field approximations} \citep{wainwright2008graphical} and \emph{normalizing flows} \citep{papamakarios2021normalizing}, offer an alternative approach. In these methods, the problem of density estimation is transformed into an optimization problem by fitting a parametric family of tractable distributions to the target density. In the context of normalizing flows, we want to mention works on constructing better loss functions~\citep{felardos2023designing} or gradient estimators~\citep{vaitl2022gradients}.

In this work, we provide a comprehensive PDE perspective on SDE-based sampling methods. Our approach is loosely inspired by~\cite{mate2023learning}, however, extended to diffusion models, optimal transport (OT), and Schrödinger bridges (SBs). Moreover, we consider other parametrizations and do not rely on the ODE for sampling the collocation points $(\xi,\tau)$. For a corresponding mean-field games (MFG) perspective, we refer to~\cite{zhang2023mean}. We also mention path space measure perspectives on SDE-based methods in~\cite{vargas2023transport,richter2023improved}.

The PDE for diffusion models has been derived in~\cite{berner2022optimal} based on prior work by~\cite{pavon1989stochastic,fleming2012deterministic} in stochastic optimal control. We refer to~\cite{chen2016relation} for the corresponding PDEs prominent in OT and SBs. Versions of the Hamilton-Jacobi-Bellman (HJB) regularizer have been used for normalizing flows in generative modeling by~\cite{onken2021ot}, for generalized SBs by~\cite{liu2022deep,koshizuka2022neural}, for MFG by~\cite{ruthotto2020machine,lin2021alternating}, and for generative adversarial models by~\cite{yang2020potential}.  

For the usage of PINNs for a generalized SB in the context of colloidal self-assembly, we refer to~\cite{nodozi2023physics}. An orthogonal direction to our approach is using divergence-free neural networks, which automatically satisfy the continuity equation and only require to fit the boundary distributions $p_\mathrm{target}$ and $p_\mathrm{prior}$~\citep{richter2022neural}. 
We further mention that higher-dimensional Fokker-Planck equations have also been tackled with time-varying Gaussian mixtures~\citep{chen2018efficient}, and there exist SDE-based neural solvers for HJB equations~\citep{richter2022robust,nusken2021solving} and combinations with PINNs~\citep{nusken2021interpolating}.

Finally, we want to highlight recent works on simulation-free learning of (stochastic) dynamics using \emph{flow matching}~\citep{tong2023improving,lipman2022flow} and \emph{action matching} techniques~\citep{neklyudov2022action}.
However, these methods rely on samples from the target distribution $p_\mathrm{target}$. Similarly, many works on solving SB and OT problems using deep learning require samples from the target distribution~\citep{chen2021likelihood,de2021diffusion,fernandes2021shooting,vargas2021solving}.

\section{Sampling via dynamical measure transport}
\label{sec: sampling via measure transport}

\begin{figure}[t]
    \centering
    \includegraphics[width=\linewidth]{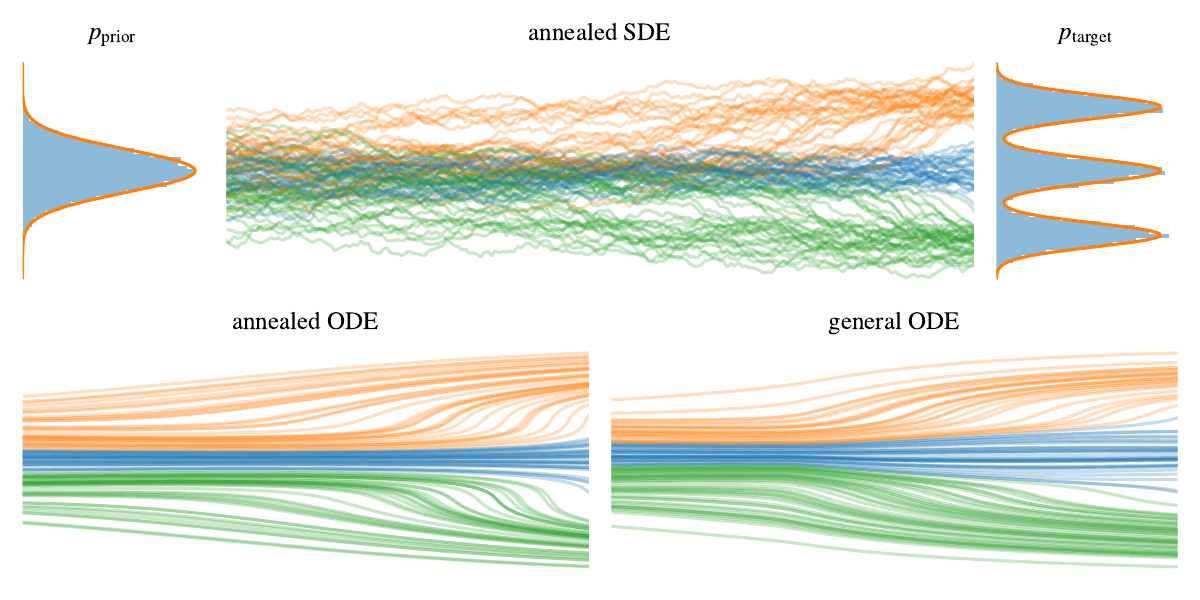}
    \caption{We plot three evolutions of the process $X$ defined in~\eqref{eq: SDE} and \eqref{eq: ODE} between a Gaussian prior density $p_{\mathrm{prior}}$ and a Gaussian mixture target density $p_{\mathrm{target}}$, corresponding to SDEs and ODEs which have been learned with three different loss functions. The top panel displays a stochastic evolution stemming from the loss $\mathcal{L}_\mathrm{logFP}^\mathrm{anneal}$, for which we additionally plot histograms of the prior and the target, respectively. In the second row we show deterministic evolutions, once obtained with $\mathcal{L}_\mathrm{logCE}^\mathrm{anneal}$ and once with $\mathcal{L}_\mathrm{logCE}$. Note that the stochastic and the left deterministic evolution follow the same annealing strategy, whereas the general loss $\mathcal{L}_\mathrm{logCE}$ leads to a different density path. We refer to \Cref{sec: learning the evolution} for the details of the different methods.}
    \label{fig:evolution}
\end{figure}

Our approach is to identify a dynamical system that transports a chosen prior density to the desired target via a deterministic or stochastic process. To be more precise, we consider the SDE
\begin{align}
\label{eq: SDE}
    \mathrm{d} X_t =  \mu(X_t, t) \, \mathrm{d}t + \sigma(t) \, \mathrm{d}W_t, \quad X_0 \sim p_\mathrm{prior},
\end{align}
where $W$ is a standard Brownian motion, or, by setting\footnote{We consider a general diffusion coefficient function $\sigma \in C( [0, T], \R^{d\times d})$, including the special case where $\sigma$ is constant zero, i.e., $\sigma=0$.} $\sigma = 0$, the ODE
\begin{align}
\label{eq: ODE}
    \mathrm{d} X_t =  \mu(X_t, t) \, \mathrm{d}t, \quad X_0 \sim p_\mathrm{prior},
\end{align}
and our goal is to learn the drift $\mu \in C(\R^d \times [0, T], \R^d)$ such that $X_T \sim p_\mathrm{target}$, see~\Cref{fig:evolution}.

Dynamical systems can be viewed on a trajectory level, as specified above, or on a density level, where we denote with $p_X(\cdot, t)$ the density of the random variable $X_t$ at time $t \in [0, T]$. It is well known that such densities can be described by PDEs \cite{pavliotis2014stochastic}
In particular, we know\footnote{We assume that the coefficient functions and densities are sufficiently regular such that we obtain unique strong solutions to the considered PDEs.} that the density $p_X$ of the stochastic process in \eqref{eq: SDE} fulfills the \textit{Fokker-Planck equation}
\begin{equation}
\label{eq: FP}
    \partial_t p_X + \div(p_X \mu) - \tfrac{1}{2} \tr(\sigma \sigma^\top \nabla^2 p_X) = 0, \quad p_X(\cdot,0)=p_{\mathrm{prior}},
\end{equation}
and, analogously, that the density $p_X$ of the deterministic process \eqref{eq: ODE} fulfills the \textit{continuity equation}
\begin{equation}
\label{eq: CE}
    \partial_t p_X + \div(p_X \mu) = 0, \quad p_X(\cdot,0)=p_{\mathrm{prior}},
\end{equation}
noting that our desired goal adds the additional boundary condition $p_X(\cdot,T)=p_{\mathrm{target}}$. A valid strategy to identify a drift that fulfills our goal is thus to look for pairs $\mu$ and $p_X$ that fulfill either of the above PDEs. It is important to note that there exist infinitely many such pairs, corresponding to infinitely many bridges between the prior and the target density. We will later discuss ways to constrain the problem, leading to unique solutions.

\vspace{0.5em}
\begin{remark}[Connections to other methods]
We note that the above framework incorporates existing sampling methods that can be related to either SDEs or ODEs. In the former setting, \emph{Schrödinger \mbox{(half-)bridges}}, \emph{diffusion models}, or \emph{annealed flows} can be understood as learning stochastic evolutions \cite{vargas2023transport,zhang2021path,berner2022optimal,richter2023improved,zhang2023diffusion,vargas2023denoising}. In the later, \emph{continuous normalizing flows} (sometimes combined with MCMC) are in instance of learned ODEs  \citep{wu2020stochastic,midgley2022flow,matthews2022continual,arbel2021annealed}. We note, however, that the previously mentioned methods rely on simulating (parts of) the process $X$ for training, which requires time discretization and typically results in unstable and slow convergence. Our PDE-based attempt, on the other hand, allows for simulation-free training, as will be explained in the next section.
\end{remark}

\section{Learning the evolution}
\label{sec: learning the evolution}

A general strategy to solve the sampling task is to identify solution pairs $\mu$ and $p_X$ that solve the PDE \eqref{eq: FP} or \eqref{eq: CE}, respectively, and our task thus corresponds to the numerical approximation of PDEs. Since our general setup allows for infinitely many solutions, it seems particularly suitable to consider variational formulations of the PDEs. To be more precise, we consider loss functionals
\begin{equation}
    \mathcal{L}: C(\R^d \times [0, T], \R^d) \times C(\R^d \times [0, T], \R) \to \R_{\ge 0},
\end{equation}
that are zero if and only if a pair $(\mu, p_X)$ fulfills the corresponding PDE. In the following, we will design different loss functions that follow the framework of PINNs, i.e. correspond to the respective PDE residual terms.

\subsection{General evolution}
\label{sec: general evolution}

Let us first study the general case. For numerical stability, it is reasonable to consider the PDEs \eqref{eq: FP} or \eqref{eq: CE} in log-space, and we note that the function $V := \log p_X$ fulfills the log-transformed Fokker-Planck equation
\begin{equation}
\label{eq: log-FK}
    \mathcal{R}_{\mathrm{logFP}}(\mu, V) := \partial_t V  + \div(\mu) + \nabla V \cdot \mu - \tfrac{1}{2}\|\sigma^\top \nabla V\|^2 - \tfrac{1}{2}\tr(\sigma \sigma^\top \nabla^2 V) = 0,
\end{equation}
or the log-transformed continuity equation
\begin{equation}
\label{eq: log-CE}
   \mathcal{R}_{\mathrm{logCE}}(\mu, V) := \partial_t V + \div(\mu) + \nabla V \cdot \mu = 0,
\end{equation}
respectively\footnote{We note that equation \eqref{eq: log-FK} is a Hamilton-Jacobi-Bellman equation when being considered as a PDE in the function $V$, see also \cite{berner2022optimal}.}. Having approximations $\widetilde{\mu}$ and $\widetilde{V}$ of the drift and log-density, we can now define losses of the type
\begin{align}
\begin{split}
\label{eq: PINN loss with boundary terms}
    \mathcal{L}(\widetilde{\mu}, \widetilde{V}) = \alpha_1 \E\left[\left(\mathcal{R}(\widetilde{\mu}, \widetilde{V})(\xi, \tau)\right)^2 \right] &+ \alpha_2 \E\left[\left(\widetilde{V}(\xi, 0) - \log p_\mathrm{prior}(\xi)\right)^2\right] \\&+ \alpha_3 \E\left[\left(\widetilde{V}(\xi, T) - \log p_\mathrm{target}(\xi)\right)^2\right], 
\end{split}
\end{align}
with suitably chosen random variables $(\xi,\tau)$ and weights $\alpha_1, \alpha_2, \alpha_3 > 0$, noting that the respective PDE is fulfilled if and only if $\mathcal{L}(\widetilde{\mu}, \widetilde{V}) = 0$.
In practice, one often chooses $(\xi,\tau) \sim \operatorname{Unif}(\Omega)$ with $\Omega \subset \R^d  \times [0, T]$ being a sufficiently\footnote{In order to solve for the exact solution, we theoretically need that the range of $(\xi,\tau)$ equals $\R^d \times [0, T]$. To mitigate a large approximation error, we thus choose a compact domain $\Omega$ large enough such that the density $p_X$ has sufficiently small values on the complement of $\Omega$.
We present further approaches in~\Cref{sec:sampling}.} large compact set. Moreover, one can consider parametrizations of the function $\widetilde{V}$ that fulfill the boundary conditions by design, e.g.,
\begin{equation}
\label{eq: V parametrization}
   \widetilde{V}_{\varphi, z}(\cdot,t) = \tfrac{t}{T} \log \tfrac{\rho_{\mathrm{target}}}{z(t)}  + \left( 1- \tfrac{t}{T}\right) \log p_{\mathrm{prior}} +  \tfrac{t}{T}\left( 1- \tfrac{t}{T}\right) \varphi(\cdot,t),
\end{equation}
such that the loss \eqref{eq: PINN loss with boundary terms} reduces to
\begin{equation}
\label{eq: PINN loss short}
    \mathcal{L}(\widetilde{\mu}, \widetilde{V}) = \E\left[\left(\mathcal{R}(\widetilde{\mu}, \widetilde{V})(\xi, \tau)\right)^2 \right],
\end{equation}
see also \cite{mate2023learning}. In the above, $z \in C([0, T], \R)$ and $\varphi \in C(\R^d \times [0, T], \R^d)$ are functions that parametrize the approximation $\widetilde{V}$. If $\varphi$ is optimized (and not fixed as, e.g., in the annealing case, see below), the function $z \in C([0, T], \R)$ can be reduced to a constant function $t \mapsto \bar{z}$, where $\bar{z} \in \R$ is a learnable parameter, see also \Cref{app: PINN loss details}. 

Specifically, we can define the two loss functions
\begin{equation}
\label{eq: PINN loss}
    \mathcal{L}_{\mathrm{logFP}}(\widetilde{\mu}, \widetilde{V}) := \E\left[\left(\mathcal{R}_{\mathrm{logFP}}(\widetilde{\mu}, \widetilde{V})(\xi, \tau)\right)^2 \right] 
\end{equation}
and
\begin{equation}
\label{eq: def L_logCE}
    \mathcal{L}_{\mathrm{logCE}}(\widetilde{\mu}, \widetilde{V}) := \E\left[\left(\mathcal{R}_{\mathrm{logCE}}(\widetilde{\mu}, \widetilde{V})(\xi, \tau)\right)^2 \right].
\end{equation}

\subsection{Constrained evolution}
\label{sec: constrained evolution}

We now discuss ways to constrain the evolution in order to get unique solutions. 
To this end, we can fix $p_X$ and only learn $\mu$ (\textit{annealing}), we can fix $\mu$ and only learn $p_X$ (\textit{time-reversal}) or we can add regularizers on $\mu$, while still learning both $\mu$ and $p_X$ (\textit{optimal transport and Schrödinger bridges}).

\textbf{Annealing.} We can prescribe a density path from prior to target by specifying $p_X$. This can, for instance, be done by choosing $\varphi=0$ in \eqref{eq: V parametrization} \cite{mate2023learning}, which yields the typical geometric path often taken in \emph{Annealed Importance Sampling} (AIS)~\cite{neal2001annealed,vargas2023transport}. This then amounts to considering the residuals
\begin{equation}
    \mathcal{R}^\mathrm{anneal}_{\mathrm{logFP}}(\widetilde{\mu}) := \mathcal{R}_{\mathrm{logFP}}(\widetilde{\mu}, V), \qquad \mathcal{R}^\mathrm{anneal}_{\mathrm{logCE}}(\widetilde{\mu}) := \mathcal{R}_{\mathrm{logCE}}(\widetilde{\mu}, V), 
\end{equation}
where now $V$ is fixed (up to the learnable normalization $z(t)$), e.g., by setting $V = V_{0, z}$ using the parametrization \eqref{eq: V parametrization}, thus yielding unique minimizers. We refer to \cite[Theorem 8.3.1]{ambrosio2005gradient}, which proves that under mild conditions we can always find a drift as the gradient of a potential, i.e. $\mu=\nabla \Phi$, such that the corresponding ODE or SDE has the prescribed density, see also \cite{neklyudov2022action}.

\textbf{Score-based generative modeling.}
For the stochastic dynamics, we may consider the concept of time-reversal as recently applied in score-based generative modeling. To this end, we may set $\mu = \sigma\sigma^\top \nabla V - f$ for a fixed function $f$, which yields
\begin{equation}
\label{eq: FK time-reversal}
    \mathcal{R}_{\mathrm{logFP}}(\sigma\sigma^\top \nabla V - f, V) := \partial_t V  - \div(f) - \nabla V \cdot f + \tfrac{1}{2}\|\sigma^\top \nabla V\|^2 + \tfrac{1}{2}\tr(\sigma \sigma^\top \nabla^2 V) = 0.
\end{equation}
One can now readily see that the time-reversal of the function $V$ fulfilling \eqref{eq: FK time-reversal}, which we denote with $\cev{V}$, fulfills (when replacing $\sigma$ with $\cev{\sigma}$)
\begin{equation}
     \mathcal{R}_{\mathrm{logFP}}(\cev{f}, \cev{V}) = 0.
\end{equation}
This corresponds to the SDE
\begin{equation}
\label{eq:inference_sde}
    \mathrm d Y_t = \cev{f}(Y_t, t) \mathrm dt + \cev{\sigma}(t) \mathrm dW_t, \qquad Y_0 \sim p_\mathrm{target},
\end{equation}
and we can thus interpret $V = \log \cev{p}_Y$, as also derived in~\cite{berner2022optimal}. In consequence, a viable strategy is to pick $f$ and $\sigma$ such that $p_Y(\cdot,T) \approx p_{\mathrm{prior}}$ (e.g., $f(x,t) = -x$ and $\sigma(t) = \sqrt{2}$, see~\Cref{sec:noise_schedule}), and minimize the loss
\begin{equation}
    \mathcal{L}_{\mathrm{score}}(\widetilde{V}) \coloneqq \mathcal{L}_{\mathrm{logFP}}(\sigma\sigma^\top \nabla \widetilde{V} - f, \widetilde{V}).
\end{equation}
For this loss, we do not need learn $z$ and enforce $\log p_{\mathrm{prior}}$ in our parametrization of $\widetilde{V}$ in~\eqref{eq: V parametrization}, since the drift $\mu$ only depends on the gradient of $\widetilde{V}$ and the boundary condition is specified by $V(\cdot,0)=\log p_Y(\cdot, T) \approx \log p_{\mathrm{prior}}$, see also~\cite{berner2022optimal}. 

\textbf{Optimal transport and Schrödinger bridges.} Another way to get unique solutions as to add a regularization to the drift. In particular, we may seek the drift $\mu$ that minimizes an energy of the form
\begin{equation}
\E \left[\tfrac{1}{2} \int_{0}^T  \|\mu(X_s,s)\|^2  \mathrm{d}s \right].
\end{equation}
For nonzero $\sigma$, this corresponds to the dynamic \emph{Schrödinger bridge} (SB) problem~\citep{dai1991stochastic}. In these cases, the optimal solution can be written as $\mu \coloneqq \nabla \Phi$,
where $\Phi$ solves the \emph{Hamilton-Jacobi-Bellman} (HJB) equation
\begin{equation}
\label{eq:hjb_pde}
    \mathcal{R}_{\mathrm{HJB}}^\mathrm{SB}(\Phi) := \partial_t \Phi + \tfrac{1}{2} \|\nabla \Phi \|^2 + \tfrac{1}{2}\tr(\sigma \sigma^\top \nabla^2 \Phi) = 0,
\end{equation}
see~\Cref{sec:hjb} and, e.g.,~\cite{pavon1991free,benamou2000computational,caluya2021wasserstein,vargas2023transport}. For $\sigma=0$, it is connected to \emph{optimal transport} (OT) problems w.r.t.\@ the Wasserstein metric~\citep{benamou2000computational} and the HJB equation turns into
\begin{equation}
\label{eq:hjb_pde OT}
         \mathcal{R}_{\mathrm{HJB}}^\mathrm{OT}(\Phi) := \partial_t \Phi + \tfrac{1}{2} \|\nabla \Phi \|^2 = 0.
\end{equation}

We can add such regularization using the losses
\begin{align}
    \label{eq:sb_loss}
   \mathcal{L}_{\mathrm{SB}}(\widetilde{\Phi}, \widetilde{V}) &:= \mathcal{L}_{\mathrm{logFP}}(\nabla \widetilde{\Phi}, \widetilde{V}) + \alpha \, \E\left[ \left( \mathcal{R}_{\mathrm{HJB}}^\mathrm{SB}(\widetilde{\Phi})(\xi,\tau) \right)^2 \right], \\
   \label{eq:ot_loss}
   \mathcal{L}_{\mathrm{OT}}(\widetilde{\Phi}, \widetilde{V}) &:= \mathcal{L}_{\mathrm{logCE}}(\nabla \widetilde{\Phi}, \widetilde{V}) + \alpha \, \E\left[ \left( \mathcal{R}_{\mathrm{HJB}}^\mathrm{OT}(\widetilde{\Phi})(\xi,\tau) \right)^2 \right],
\end{align}
where $\alpha > 0$ is a suitably chosen weight.

\subsection{Connections to previous attempts}

\begin{table*}[t]
\renewcommand{\arraystretch}{1.2}
\centering
\caption{Summary of our considered losses. Empty cells do not have a direct correspondence.}
\resizebox{\textwidth}{!}{
\begin{tabular}{ l c c c c }
  \toprule
 Method & Stochastic & Deterministic & BSDE version & Unique \\ 
  \midrule
 General bridge & $\mathcal{L}_\mathrm{logFP}(\widetilde{\mu}, \widetilde{V})$ & $\mathcal{L}_\mathrm{logCE}(\widetilde{\mu}, \widetilde{V})$ & Bridge~\cite{richter2023improved,chen2021likelihood}  & \xmark \\
 Prescribed / annealed bridge &$\mathcal{L}^\mathrm{anneal}_\mathrm{logFP}(\widetilde{\mu})$ & $\mathcal{L}^\mathrm{anneal}_\mathrm{logCE}(\widetilde{\mu})$ & CMCD~\cite{vargas2023transport} & \cmark \\ 
Time-reversal / diffusion model &$\mathcal{L}_\mathrm{score}(\widetilde{V})$ &  & DIS~\cite{berner2022optimal} & \cmark \\ 
Regularized drift / SB / OT &$\mathcal{L}_\mathrm{SB}(\widetilde{\mu}, \widetilde{V})$ & $\mathcal{L}_\mathrm{OT}(\widetilde{\mu}, \widetilde{V})$ & &  \cmark \\
 \bottomrule
\end{tabular}}
\label{tab: losses}
\end{table*}

In this section, we show that we can re-derive already existing methods in diffusion-based sampling via our PDE perspective. This can be done by replacing our PINN-based losses with losses based on backward stochastic differential equations (BSDEs). Those losses build on a stochastic representation of the PDE at hand, essentially coming from It\^{o}'s formula, see \cite{nusken2021interpolating} and the references therein for details. In the following proposition we relate BSDE-based versions of our losses to alternative trajectory-based losses, indicated by $\mathcal{L}_\mathrm{method}^\mathrm{BSDE}$. We provide an overview in~\Cref{tab: losses}. Interestingly, the proposition shows that many of the diffusion-based methods can in fact be derived without the concept of time-reversal. We refer to \Cref{app: BSDE-based losses} for the proof and further details.

\begin{proposition}[Equivalence to trajectory-based methods]
\label{prop: BSDE loss equivalences}
The BSDE versions of our losses are equivalent to previously existing losses in the following sense.
\begin{enumerate}[label=(\roman*)]
    \item Assuming the reparametrization $\widetilde{\mu} = f + \sigma u$ and $\sigma^\top \nabla \widetilde{V} = u + v$, it holds
    \begin{equation}
        \mathcal{L}^{\mathrm{BSDE}}_{\mathrm{logFP}}(\widetilde{\mu},\widetilde{V}) = \mathcal{L}^{\mathrm{BSDE}}_{\mathrm{Bridge}}(u, v),
    \end{equation} 
    where $\mathcal{L}^{\mathrm{BSDE}}_{\mathrm{Bridge}}$ is derived in \cite{richter2023improved}.
    \item It holds 
    \begin{equation}
    \mathcal{L}_{\mathrm{logFP}}^\mathrm{anneal,BSDE}(\widetilde{\mu}) = \mathcal{L}_{\mathrm{CMCD}}^\mathrm{BSDE}(\widetilde{\mu}),
    \end{equation}
    where $\mathcal{L}_{\mathrm{CMCD}}^\mathrm{BSDE}$ refers to (a version of) the \textit{Controlled Monte Carlo Diffusion} (CMCD) loss derived in \cite{vargas2023transport}.
    \item Assuming the reparametrization $\widetilde{\mu} = f + \sigma u$ and $\sigma^\top \nabla \widetilde{V} = u$, it holds
    \begin{equation}
     \mathcal{L}_\mathrm{score}^\mathrm{BSDE}(\widetilde{V}) = \mathcal{L}_\mathrm{DIS}^\mathrm{BSDE}(u),
    \end{equation}
    where $\mathcal{L}_\mathrm{DIS}^\mathrm{BSDE}$ refers to the \textit{Time-Reversed Diffusion Sampler} (DIS) loss derived in \cite{berner2022optimal}.
\end{enumerate}
\end{proposition}

\vspace{0.5em}
\begin{remark}[Numerical implications of PINN- and BSDE-based losses]
From a numerical perspective, the derived PINN- and BSDE-based losses have advantages and disadvantages. Since BSDE-based methods build on a stochastic representation of the PDE, neither second-order nor time derivatives have to be computed. This also leads to the fact that for our sampling problems, the gradients of the solutions (usually corresponding to the learned drift) can be learned directly. It comes at the price, however, that only stochastic dynamics can be approached. PINN-based losses, on the other hand, are more general, e.g., they can be readily applied to deterministic evolutions as well. Moreover, they are simulation-free and do not rely on time-discretization, overall resulting in lower times per gradient steps for moderate dimensions. Furthermore, off-policy training basically comes by design, which might be advantageous for mode discovery.
\end{remark}

\vspace{0.5em}
\begin{remark}[Subtrajectory-based losses]
Another equivalence can be deduced when considering the \textit{diffusion loss} introduced in \cite{nusken2021interpolating} instead of the BSDE loss, which does not aim to learn It\^{o}'s formula on the entire time interval, but rather on subintervals $[t_0, t_1] \subset [0, T]$, which may be randomly drawn during optimization. Along the lines of \Cref{prop: BSDE loss equivalences}, one can readily show that applying the \textit{diffusion loss} to the log-transformed Fokker-Planck equation \eqref{eq: log-FK}, one can recover subtrajectory-based losses suggested e.g. in \cite{zhang2023diffusion}, see also \cite[Appendix A.7]{richter2023improved}.
\end{remark}

\section{Gauss-Newton methods for improved convergence of PINNs}
\label{sec:gn}
Training physics-informed neural networks can be challenging. It is well-documented in the literature that differential operators in the loss function complicate the training and can lead to ill-conditioning \cite{de2023operator,  krishnapriyan2021characterizing, wang2021understanding}. At the same time, accurate solutions are crucial for achieving high sampling quality. To obtain optimal results in PINN training, we therefore combine the Adam optimizer with a Gauss-Newton method which we derive from an infinite-dimensional function space perspective. This viewpoint has recently been explored in \cite{muller2023achieving, muller2024optimization}.

\textbf{Gauss-Newton method in function space.} We consider loss functions of the form 
\begin{equation}\label{eq:nonlinear_ls}
    \mathcal{L}(\widetilde{V}) 
    \coloneqq
    \E\left[ \left( \mathcal{R}(\widetilde{V})(\xi, \tau) \right)^2 \right],
\end{equation}
where $\mathcal{R}$ is a nonlinear PDE operator. For example, derived from \eqref{eq: FK time-reversal}, we set $\mathcal R = \mathcal R_{\mathrm{score}}$ to be\footnote{Here, we assumed that initial and final conditions are exactly satisfied, as described in Section~\ref{sec: general evolution}.}
\begin{equation*}
    \mathcal{R}_{\mathrm{score}}(\widetilde{V})
    =
    \partial_t \widetilde{V}  
    -
    \operatorname{div}(f)
    -
    \nabla \widetilde{V}\cdot f
    +
    \tfrac12 \|\sigma^\top\nabla \widetilde{V} \|^2 
    +
    \tfrac12\tr(\sigma \sigma^\top \nabla^2 \widetilde{V}).
\end{equation*}
To optimize $\mathcal L$ \emph{in function space}, a sensible choice is Gauss-Newton's method for nonlinear least-squares problems, due to its local quadratic convergence properties \cite{deuflhard1979affine} and documented success in PINN training \cite{jnini2024gauss}, which are to be contrasted to much slower rates of first-order methods like gradient descent \cite{nocedal1999numerical}. The rationale of Gauss-Newton is to linearize $\mathcal{R}$ in the least squares formulation \eqref{eq:nonlinear_ls} and to solve the resulting quadratic minimization problem at every step. More precisely, choosing a start value $\widetilde{V}_0$, we optimize $\mathcal L$ via
\begin{equation}\label{eq:f_space_gn}
    \widetilde{V}_{k+1}
    =
    \widetilde{V}_k
    -
    [D\mathcal{R}_{\mathrm{score}}(\widetilde{V}_k)^*D \mathcal{R}_{\mathrm{score}}(\widetilde{V}_k)]^{-1}(D\mathcal L(\widetilde{V}_k)),
    \quad 
    k=0,1,2,\dots
\end{equation}
where $D\mathcal L$ and $D\mathcal R$ denote the Fr\'echet derivatives of $\mathcal L$ and $\mathcal R$, respectively, and $D\mathcal R(\widetilde{V}_k)^*$ is the adjoint of $D\mathcal R(\widetilde{V}_k)$.
In the case of the example above, i.e., equation \eqref{eq: FK time-reversal},
it holds 
\begin{equation}
    D \mathcal{R}_{\mathrm{score}}(\widetilde{V})[\delta_{\widetilde{V}}] 
    =
    \partial_t \delta_{\widetilde{V}} 
    -
    \nabla \delta_{\widetilde{V}}\cdot f
    +
    \sigma^\top\nabla \widetilde{V} \cdot \nabla \delta_{\widetilde{V}} 
    +
    \tr(\sigma\sigma^\top \nabla^2 \delta_{\widetilde{V}})
\end{equation}
and the computation of the inverse entails solving the following PDE at every step of the iteration: Find $\delta_{\widetilde{V}}$ such that for all $\bar\delta_{\widetilde{V}}$ (in a suitable test space) it holds
\begin{equation*}
    \E\left[
    D \mathcal{R}_{\mathrm{score}}(\widetilde{V}_k)[\delta_{\widetilde{V}}](\xi, \tau)
    D \mathcal{R}_{\mathrm{score}}(\widetilde{V}_k)[\bar\delta_{\widetilde{V}}](\xi, \tau)
    \right]
    =
    D\mathcal L({\widetilde{V}}_k)(\bar \delta_{\widetilde{V}}).
\end{equation*} 
To transfer this function space optimization to a computable algorithm for neural network optimization, we discretize it in the tangent space of the neural network ansatz. The advantage of this approach is that we are guaranteed to follow the dynamics of \eqref{eq:f_space_gn} up to a projection onto the tangent space \cite[Theorem 1]{muller2024optimization}. To make the dependence of a neural network approximation $\widetilde V$ on the trainable parameters explicit, we write $\widetilde V = V_\theta$. Here, the vector $\theta\in\mathbb R^p$ collects the $p$ trainable parameters of the neural network ansatz. Discretizing the algorithm~\eqref{eq:f_space_gn}, we obtain an iteration of the form
\begin{equation*}
    \theta_{k+1} = \theta_k - \eta_k G(\theta_k)^\dagger \nabla L(\theta_k), \quad k=0,1,2,\dots
\end{equation*}
where $L(\theta) = \mathcal L(V_\theta)$ and $\nabla L(\theta)$ denotes the gradient of $L$ w.r.t. $\theta$, typically computed via automatic differentiation. By $\eta_k>0$ we denote a step-size and $G(\theta_k)^\dagger$ is the Moore-Penrose inverse of the Gramian $G(\theta_k)$. The matrix $G(\theta_k)$ is derived from the operator $D \mathcal{R}_{\mathrm{score}}(V_{\theta_k})^*D\mathcal{R}_{\mathrm{score}}(V_{\theta_k})$ via
\begin{equation*}
    G(\theta_k)_{ij}
    =
    \E\left[
    D\mathcal{R}_{\mathrm{score}}(V_{\theta_k})[\partial_{\theta_i}V_{\theta_k}](\xi, \tau)
    D\mathcal{R}_{\mathrm{score}}(V_{\theta_k})[\partial_{\theta_j}V_{\theta_k}](\xi, \tau)
    \right].
\end{equation*}
It is detailed in \cite[Appendix C]{muller2024optimization} that this approach corresponds to the standard Gauss-Newton method for a suitably chosen residual. As a standard Gauss-Newton method, it can be implemented in a matrix-free way \cite{schraudolph2002fast}, relying on an iterative solver, such as the conjugate gradient method to obtain $G(\theta_k)^\dagger \nabla L(\theta)$. In practice, we use an additive damping, i.e., we use $G(\theta_k) + \varepsilon\mathrm{I}$ instead of $G(\theta_k)$, for some $\varepsilon>0$, which guarantees invertibility of the matrix.

\section{Numerical experiments}
\label{sec: numerics}

In this section, we evaluate our PINN-based losses on different benchmark problems. Specifically, we consider the losses listed in \Cref{tab: losses} and compare them with state-of-the-art trajectory-based methods. For the benchmark problems, we follow~\cite{richter2023improved} and consider a Gaussian mixture model as well as high-dimensional, multimodal many-well distributions, which resemble typical problems in molecular dynamics. We refer to~\Cref{sec:app_exp} for a description of the targets and details on our implementation. In our experiments, we compare against the \emph{Path Integral Sampler} (PIS)~\citep{zhang2021path} and the \emph{Time-Reversed Diffusion Sampler} (DIS)~\citep{berner2022optimal}, including the log-variance loss suggested in~\cite{richter2023improved}. 

\begin{figure}[t]
    \centering
    \begin{minipage}{0.5\linewidth}
        \centering
        \includegraphics[width=0.95\linewidth]{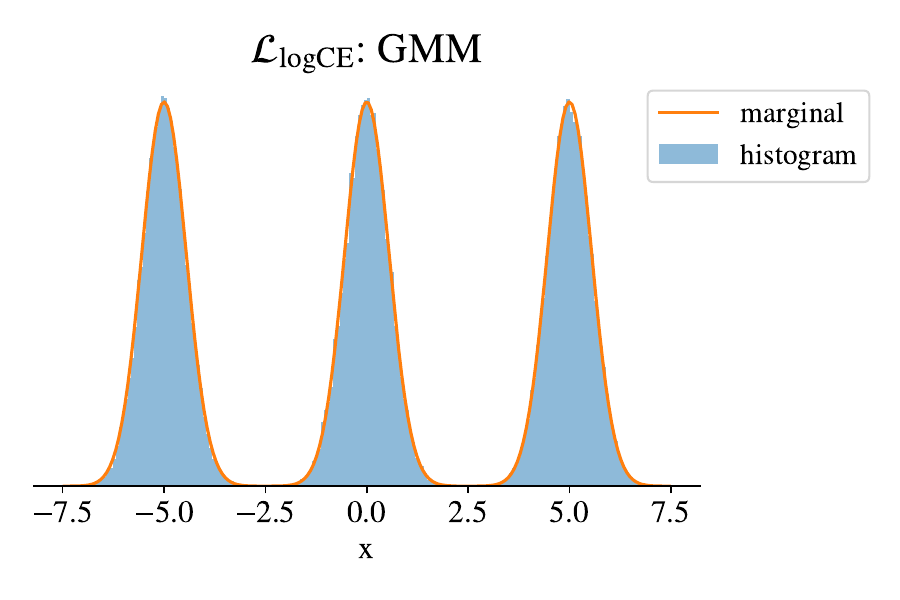}
    \end{minipage}%
    \begin{minipage}{0.5\linewidth}
        \centering
        \includegraphics[width=0.95\linewidth]{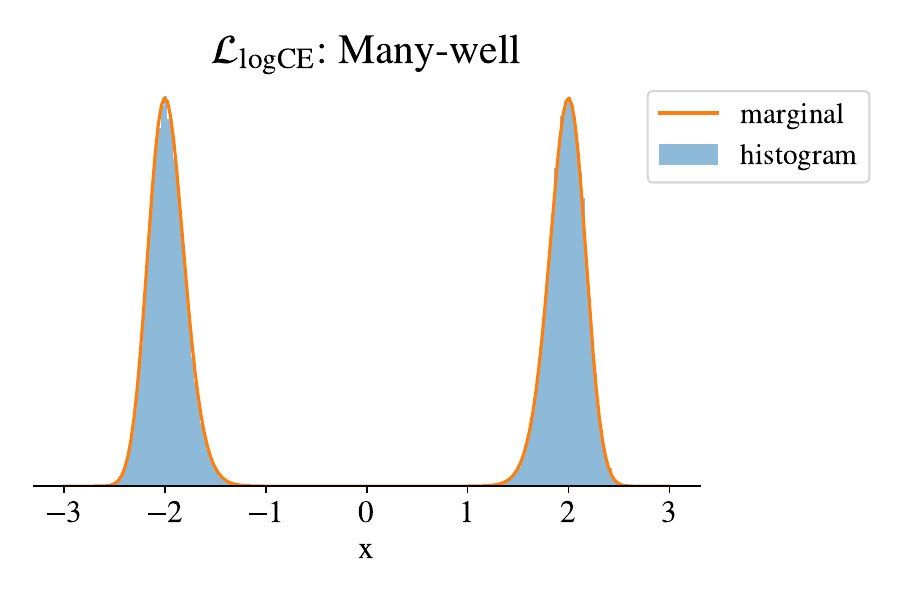}
    \end{minipage}
    \vspace{-0.75em}
    \caption{The ground truth marginal in the first dimension and histograms of samples from our best performing method using the loss $\mathcal{L}_\mathrm{logCE}$ on the GMM (left) and many-well (right) examples.}
    \label{fig:marginals}
\end{figure}

Our results are summarized in~\Cref{tab:results} -- in order to have a fair comparison with PIS and DIS we did not employ the Gauss-Newton method here. In general, we see that the ODE methods usually outperform the SDE methods, in particular significantly improving upon the baseline methods. In~\Cref{fig:marginals} we illustrate that we can indeed accurately cover the modes of the target distributions. We refer to~\Cref{fig:marg_sde,fig:marg_ode} in the appendix for additional visualizations. In \Cref{tab:results_gn}, we report results attained by fine-tuning with the Gauss-Newton method derived in~\Cref{sec:gn}, showing that we can indeed further improve sampling performance.

In general, we also observe that learning a potential $\Phi$ (with $\mu=\nabla \Phi$) rather than the drift $\mu$ directly, such as in the SB and OT losses, is more challenging and can lead to worse performance. In particular, the HJB regularization only provides good results for the ODE case. Interestingly, there is no clear advantage of the methods with prescribed density (i.e., using the losses $\mathcal{L}_\mathrm{logFP}^\mathrm{anneal}, \mathcal{L}_\mathrm{logCE}^\mathrm{anneal}, \mathcal{L}_\mathrm{score}$), indicating that, in general, non-uniqueness might improve performance. For the annealing losses, we presume that the performance significantly depends on the chosen annealing strategies. In particular, the applied geometric annealing defined in \Cref{sec: constrained evolution} is known to be suboptimal for certain prior-target configurations, see, e.g., \cite{neal2001annealed,gelman1998simulating} and \Cref{app: annealing challenges} for an illustrative example. 

\begin{table*}[!t]
\renewcommand{\arraystretch}{1.2}
\setlength{\aboverulesep}{0pt}
\setlength{\belowrulesep}{0pt}
\centering
\caption{Metrics for the benchmark problems in different dimensions $d$. We report errors for estimating the log-normalizing constant ($\Delta \log Z$) and the standard deviations of the marginals ($\Delta \operatorname{std}$). Furthermore, we report the normalized effective sample size ($\operatorname{ESS}$) and the Sinkhorn distance ($\mathcal{W}^2_\gamma$)~\citep{cuturi2013sinkhorn}, see~\Cref{app: metrics} for details. Finally, we present the time in seconds for one gradient step. The arrows $\uparrow$ and $\downarrow$ indicate whether we want to maximize or minimize a given metric. Our methods are colored in blue (SDE) and dark blue (ODE).}
\resizebox{\textwidth}{!}{
\begin{tabular}{lllrrrrr}
    \toprule
     Problem & Method& Loss  & \hspace{-1em} $\Delta\log Z\downarrow$ & \hspace{0.05em} $\mathcal{W}^2_\gamma \downarrow$ & \hspace{0.2em} $\operatorname{ESS} \uparrow$ & \hspace{-0.8em} $\Delta\operatorname{std} \downarrow$  & \hspace{-0.5em}sec./it. $\downarrow$ \\
    \midrule
    GMM
    & PIS-KL~\citep{zhang2021path} & & 1.094 & 0.467 & 0.0051 & 1.937 & 0.503 \\
    {\scriptsize $(d=2)$}  & PIS-LV~\citep{richter2023improved} && 0.046 & \textbf{0.020} & 0.9093 & 0.023 & 0.500 \\
    & DIS-KL~\citep{berner2022optimal} & & 1.551 & 0.064 & 0.0226 & 2.522 & 0.565 \\
    & DIS-LV~\cite{richter2023improved} & & 0.056 & \textbf{0.020} & 0.8660 & 0.004 & 0.536 \\
   \cmidrule{2-8}
    \rowcolor{blue!2.5}
    & SDE&  $\mathcal{L}_\mathrm{logFP} $ & \textbf{0.000} &    \textbf{0.020} &               \textbf{1.0000}            &        0.004    & 0.011  \\
    \rowcolor{blue!2.5}
    & SDE-anneal& $\mathcal{L}_\mathrm{logFP}^\mathrm{anneal} $ & 5.364 &             0.172 &                0.1031            &         0.209  & 0.062   \\
    \rowcolor{blue!2.5}
    & SDE-score& $\mathcal{L}_\mathrm{score} $  & 0.009 &             \textbf{0.020} &             0.9818           &   0.096   & 0.013   \\
    \rowcolor{blue!2.5}
    & SB& $\mathcal{L}_\mathrm{SB} $ & 0.002 &             \textbf{0.020} &            0.9959              &       0.050    & 0.017    \\
   \cmidrule{2-8}
    \rowcolor{blue!5}
    & ODE& $\mathcal{L}_\mathrm{logCE} $ & \textbf{0.000} &             
    \textbf{0.020} &                  \textbf{1.0000}          &       \textbf{0.003}   & \textbf{0.008}   \\
    \rowcolor{blue!5}
    & ODE-anneal& $\mathcal{L}_\mathrm{logCE}^\mathrm{anneal} $ & 4.227 &             0.044 &              0.1427             &   0.753 & 0.020  \\
    \rowcolor{blue!5}
    & OT& $\mathcal{L}_\mathrm{OT} $ & 0.005 &             0.057 &                0.9932            &         0.065  & 0.080   \\
    
    \midrule
    MW & PIS-KL~\citep{zhang2021path} &&  	3.567 & 1.699 & 0.0004 & 1.409 & 0.441 \\
    {\scriptsize $(d=5,m=5,\delta=4)$} \hspace{-0.4em} & PIS-LV~\citep{richter2023improved}& & 0.214 & 0.121 & 0.6744 & 0.001 & 0.402 \\
    
    & DIS-KL~\citep{berner2022optimal} && 1.462 & 1.175 & 0.0012 & 0.431 & 0.490 \\
    & DIS-LV~\citep{richter2023improved}& & 0.375 & 0.120 & 0.4519 & 0.001 & 0.437 \\

   \cmidrule{2-8}
    \rowcolor{blue!2.5}
    & SDE & $\mathcal{L}_\mathrm{logFP}$ &  0.161  &             0.123 &            0.8167               &    0.016     & 0.017     \\
    \rowcolor{blue!2.5}
    & SDE-anneal& $\mathcal{L}_\mathrm{logFP}^\mathrm{anneal} $ &  0.842 &             0.257 &             0.3464               &      0.004   & 0.014   \\
    \rowcolor{blue!2.5}
    & SDE-score& $\mathcal{L}_\mathrm{score} $ &  3.969 &             0.427 &             0.0124               &      0.004   & 0.026   \\
    \rowcolor{blue!2.5}
    
    & SB & $\mathcal{L}_\mathrm{SB} $ & 7.855 &             0.328 &                0.0314           &      0.045    & 0.029    \\
   \cmidrule{2-8}
    \rowcolor{blue!5}
    & ODE & $\mathcal{L}_\mathrm{logCE} $ & \textbf{0.000} &             \textbf{0.118} &            \textbf{0.9993}               &       \textbf{0.000}    & \textbf{0.008}   \\
    \rowcolor{blue!5}
    & ODE-anneal& $\mathcal{L}_\mathrm{logCE}^\mathrm{anneal} $ & 0.025 &             0.121 &                           0.9506 &             0.005 & 0.010\\
    \rowcolor{blue!5}
    & OT & $\mathcal{L}_\mathrm{OT} $ & 0.010 &             0.120 &             0.9862             &        0.002  & 0.020    \\
    \midrule
    MW & PIS-KL~\citep{zhang2021path}& & 0.101 & 6.821 & 0.8172 & 0.001 & 0.479 \\
    {\scriptsize $(d=50,m=5,\delta=2)$} \hspace{-0.4em} & PIS-LV~\citep{richter2023improved}& & 0.087 & 6.823 & 0.8453 & \textbf{0.000} & 0.416 \\
     & DIS-KL~\citep{berner2022optimal}& & 1.785 & 6.854 & 0.0225 & 0.009 & 0.522 \\
    & DIS-LV~\citep{richter2023improved}& & 1.783 & 6.855 & 0.0227 & 0.009 & 0.450 \\
   \cmidrule{2-8}
    \rowcolor{blue!2.5}
    & SDE  &$\mathcal{L}_\mathrm{logFP} $& 0.038 &             6.820 &                           0.9511 &             0.001 & 0.050 \\
    \rowcolor{blue!2.5}
    & SDE-anneal &$\mathcal{L}_\mathrm{logFP}^\mathrm{anneal} $ & 0.270 &             6.899 &                           0.9171 &             0.021 & 0.067 \\
    \rowcolor{blue!2.5}
    & SDE-score &$\mathcal{L}_\mathrm{score} $ & 1.989 &             \textbf{6.803} &            0.1065                & 0.016   & 0.053         \\
    \rowcolor{blue!2.5}
    & SB &$\mathcal{L}_\mathrm{SB} $ & 189.71 &             7.552 &                           0.0106 &             0.051  & 0.053\\
   \cmidrule{2-8}
    \rowcolor{blue!5}
    & ODE &$\mathcal{L}_\mathrm{logCE} $ & \textbf{0.003} &             6.815 &                           \textbf{0.9937} &             0.002 & \textbf{0.023} \\
    \rowcolor{blue!5}
    & ODE-anneal &$\mathcal{L}_\mathrm{logCE}^\mathrm{anneal} $ & 1.759 &             6.821 &            0.2100               &     0.017  & 0.043            \\
    \rowcolor{blue!5}
    & OT& $\mathcal{L}_\mathrm{OT} $ & 0.104 &             6.824 &                           0.9027 &             0.001 & 0.043\\
    \bottomrule
\end{tabular}}
\vspace{-0.5em}
\label{tab:results}
\end{table*}

\begin{table*}[!t]
\renewcommand{\arraystretch}{1.2}
\setlength{\aboverulesep}{0pt}
\setlength{\belowrulesep}{0pt}
\centering
\caption{Comparison of training only with Adam ($200k$ iterations) versus pretraining with Adam ($100k$ iterations) and finetuning with the Gauss-Newton (GN) method ($500$ iterations) for the loss $\mathcal{L}_\mathrm{logCE}$. For the GN method, we use a maximum of $500$ steps for the conjugate gradient method, a damping of $\varepsilon=10^{-5}$, and a line-search for the learning rate. We refer to~\Cref{app: metrics} and~\Cref{tab:results} for details on the metrics. The arrows $\uparrow$ and $\downarrow$ indicate whether we want to maximize or minimize a given metric.}

\resizebox{\textwidth}{!}{
\begin{tabular}{llrrrrrr}
    \toprule
     Problem & Optimizer  & Loss $\downarrow$ & \hspace{-0.5em} $\Delta\log Z\downarrow$ & \hspace{0.05em} $\mathcal{W}^2_\gamma \downarrow$ & \hspace{-0.4em} $1-\operatorname{ESS} \downarrow$ & \hspace{-0.8em} $\Delta\operatorname{std} \downarrow$ & sec./it. $\downarrow$ \\ 
    \midrule
    GMM & Adam & 4.62e-4 & 3.73e-5 & 2.03e-2 & 3.15e-5 & 3.16e-3 & \textbf{0.007} \\
    {\scriptsize $(d=2)$} & Adam+GN & \textbf{1.62e-4} & \textbf{2.91e-6} & \textbf{2.03e-2} & \textbf{7.23e-6} & \textbf{1.33e-3} & 6.071  \\  
    \midrule
    MW    & Adam & 3.27e-3 & \textbf{8.79e-5} & 1.18e-1 &    6.62e-4 &       3.06e-4 & \textbf{0.008} \\
    {\scriptsize $(d=5,m=5,\delta=4)$} & Adam+GN & \textbf{2.57e-3} & 2.15e-4 & \textbf{1.18e-1} & \textbf{1.56e-4} & \textbf{1.32e-4} & 7.486 \\ 
    \bottomrule
\end{tabular}}
\label{tab:results_gn}
\end{table*}

\subsection{Limitations}
\label{sec: limitations}

We note that our approach assumes knowledge of a suitable set $\Omega\subset \R^d \times [0, T]$ for sampling the random variable $\xi$, i.e. the data on which the PINN loss is evaluated. We incur an approximation error if the set $\Omega$ is chosen too small. On the other hand, if it is too large, low probability areas of $p_{\mathrm{target}}$ can lead to instabilities and might require clipping. We provide initial results for adaptive methods in~\Cref{sec:app_exp} and leave an extensive evaluation for future work. We also mention that the computation of divergences and Laplacians using automatic differentiation can be prohibitive in very high dimensions and might require (stochastic) estimators, such as Hutchinson's trace estimator. Finally, it is commonly known that PINNs can be sensitive to hyperparameter settings.

\section{Conclusion}

We provide a principled framework for dynamical measure transport based on SDEs that allows the use of PINNs for sampling from unnormalized densities. In particular, the framework allows us to learn the drifts of SDEs or ODEs in order to end up at the target density in a finite time. The PDE framework unifies various sampling methods that are based on, e.g., normalizing flows, diffusion models, optimal transport, and Schrödinger bridges, but also adds novel approaches, e.g., by accepting non-unique solutions. Moreover, it yields flexible objectives that are free of time-discretizations and simulations. We benchmark our methods on multimodal target distributions with up to $50$ dimensions. While some SDE-based methods are still unstable, ODE-based variants yield competitive methods that can outperform various baselines. We anticipate that our methods can be improved even further using combinations with simulation-based losses as well as common tricks for PINNs, see~\Cref{sec:app_exp}. 

\subsection*{Acknowledgements}
The research of L. Richter\@ was partially funded by Deutsche Forschungsgemeinschaft (DFG) through the grant CRC 1114 ``Scaling Cascades in Complex Systems'' (project A05, project number 235221301). J. Berner acknowledges support from the Wally Baer and Jeri Weiss Postdoctoral Fellowship. A. Anandkumar is supported in part by Bren endowed chair and by the AI2050 senior fellow program at Schmidt Sciences.

\clearpage

\bibliographystyle{plain}
\bibliography{bib}

\clearpage

\appendix

\section{Theoretical aspects}
\subsection{Details on PINN-based losses}
\label{app: PINN loss details}

In this section, we will elaborate on details regarding the PINN-based losses introduced in \Cref{sec: learning the evolution}.
We first remark that under mild conditions the Fokker-Planck and continuity equations in~\eqref{eq: FP} and~\eqref{eq: CE} are mass conserving, i.e., $\partial_t \int p_X(x,t)\, \mathrm{d}x = 0$ for $t \in [0,T]$. In particular, since our initial condition $p_{\mathrm{prior}}$ is normalized, the solution $p_X(\cdot, t)$ needs to also be normalized for all $t \in [0,T]$. We thus need to make our parametrization \eqref{eq: V parametrization}, i.e.,
\begin{equation}
   \widetilde{V}_{\varphi, z}(\cdot,t) = \tfrac{t}{T} \log \tfrac{\rho_{\mathrm{target}}}{z(t)}  + \left( 1- \tfrac{t}{T}\right) \log p_{\mathrm{prior}} +  \tfrac{t}{T}\left( 1- \tfrac{t}{T}\right) \varphi(\cdot,t),
\end{equation}
sufficiently expressive. For the annealing case (i.e., when considering the losses $\mathcal{L}_\mathrm{logFP}^\mathrm{anneal}$ or $\mathcal{L}_\mathrm{logCE}^\mathrm{anneal}$), we therefore need to use a time-dependent function $z \in C([0,T], \R)$ (as opposed to a constant), since otherwise $ p_X = \exp(\widetilde{V})$ could, in general, not be a normalized density for $t \in (0, T)$. Note that if $\widetilde{V}_{\varphi, z} = V$, i.e. if it fulfills the log-transformed Fokker-Planck equation \eqref{eq: log-FK}, conservation of mass implies that $z(T)=Z$ and thus the terminal condition $\widetilde{V}_{\varphi, z}(\cdot,T) = \log p_{\mathrm{target}}$ is satisfied.

\subsection{Schrödinger bridges and Hamilton-Jacobi-Bellman equation}
\label{sec:hjb}

Let us present a sketch of the proof that the optimal drift for a prescribed density can be written as a gradient field, which, in the case of  Schrödinger bridge or optimal transport problems, solves an HJB equation, see also~\cite{pavon1991free,neklyudov2022action,koshizuka2022neural,benamou2000computational,caluya2021wasserstein}.
Let us consider the optimization problem
\begin{subequations}
\begin{align}
    \label{eq:cost}
    &\inf_{\mu} \ \frac{1}{2} \int_{0}^T \int_{\R^d} \|\mu(x,t)\|^2 \,  p(x,t) \, \mathrm{d}x \, \mathrm{d}t \\
    &\text{s.t.} \quad \partial_t p = -\div(p \mu) + \frac{1}{2} \tr(\sigma \sigma^\top \nabla^2 p), \quad p(\cdot,0)=p_{\mathrm{prior}} \quad p(\cdot,T)=p_{\mathrm{target}},
\end{align}
\end{subequations}
for a sufficiently smooth density $p$. Introducing a Lagrange multiplier $\Phi\colon \R^d \times [0,T]\to \R$, we can rewrite the problem as
\begin{align}
\label{eq:lagrange}
    &\sup_{\Phi} \inf_{\mu} \  \int_{0}^T \int_{\R^d}\,\left( \frac{1}{2}\|\mu\|^2 \,  p + \Phi\left( \partial_t p + \div(p \mu) - \frac{1}{2} \tr(\sigma \sigma^\top \nabla^2 p) \right) \right) \mathrm{d}x \, \mathrm{d}t, 
\end{align}
where we omit here and in the following the arguments of the functions for notational convenience. Using integration by parts, we can calculate
\begin{equation}
    \int_{0}^T  \Phi \partial_t p \, \mathrm{d}t = \big[\Phi p \big]_{t=0}^{t=T} - \int_{0}^T  p \, \partial_t \Phi \, \mathrm{d}t.
\end{equation}
and
\begin{equation}
    \int_{\R^d} \, \Phi \tr(\sigma \sigma^\top \nabla^2 p) \, \mathrm{d}x =  \int_{\R^d} \,  p \tr(\sigma \sigma^\top \nabla^2 \Phi) \, \mathrm{d}x,
\end{equation}
where we assume that $p$ and its partial derivatives vanish sufficiently fast at infinity.
Using the product rule and Stokes' theorem, we obtain that
\begin{equation}
    \int_{\R^d}\, \Phi \div(p \mu)\, \mathrm{d}x = \int_{\R^d}\,  \div( \Phi p \mu)\, \mathrm{d}x - \int_{\R^d}\,  p\, \mu \cdot  \nabla \Phi  \, \mathrm{d}x = - \int_{\R^d}\,  p\, \mu \cdot  \nabla \Phi  \, \mathrm{d}x.
\end{equation}
Leveraging Fubini's theorem and combining the last three calculations with~\eqref{eq:lagrange}, we obtain that
\begin{align}
\label{eq:lagrange_rewrite}
    &\sup_{\Phi} \inf_{\mu} \  \int_{\R^d} \int_{0}^T \left(\left( \left( \frac{1}{2}\|\mu\|^2  - \mu \cdot  \nabla \Phi \right) p  -   \left(\partial_t \Phi   + \frac{1}{2}  \tr(\sigma \sigma^\top \nabla^2 \Phi) \right)p  \right) \mathrm{d}t +  \big[\Phi p \big]_{t=0}^{t=T} \right) \mathrm{d}x.   
\end{align}
In view of the binomial formula, we observe that the minimizer is given by 
\begin{equation}
    \mu = \nabla \Phi.
\end{equation}
We can thus write~\eqref{eq:lagrange_rewrite} as
\begin{align}
    &\inf_{\Phi}  \  \int_{\R^d}  \int_{0}^T \left( \left( \partial_t \Phi  + \frac{1}{2}\|\nabla \Phi \|^2   + \frac{1}{2} \tr(\sigma \sigma^\top \nabla^2\Phi) \right)p \, \mathrm{d}t -  \big[\Phi p \big]_{t=0}^{t=T} \right) \, \mathrm{d}x,
\end{align}
which corresponds to the action matching objective in~\cite{neklyudov2022action}. We also refer to \cite[Theorem 8.3.1]{ambrosio2005gradient} for existence and uniqueness results. If we additionally minimize~\eqref{eq:cost} over all densities $p$ with $p(\cdot,0)=p_{\mathrm{prior}}$ and $p(\cdot,T)=p_{\mathrm{target}}$, we obtain the problem
\begin{align}
    &\inf_{\Phi,p}  \  \int_{\R^d}  \int_{0}^T \left(  \left( \partial_t \Phi  + \frac{1}{2}\|\nabla \Phi \|^2 + \frac{1}{2} \tr(\sigma \sigma^\top \nabla^2\Phi)  \right)p \, \mathrm{d}t -  \big[\Phi p \big]_{t=0}^{t=T}  \right) \mathrm{d}x,
\end{align}
Computing the functional derivative w.r.t.\@ $p$, we obtain the first-order optimality condition
\begin{equation}
    \partial_t \Phi  = - \frac{1}{2} \tr(\sigma \sigma^\top \nabla^2\Phi) - \frac{1}{2}\|\nabla \Phi \|^2,  
\end{equation}
which yields the Hamilton-Jacobi-Bellman equation in~\eqref{eq:hjb_pde}.

\subsection{BSDE-based losses and equivalences with diffusion-based sampling methods}
\label{app: BSDE-based losses}

In this chapter, we give some background on BSDE-based losses for PDEs and will show that with our PDE framework we can recover already existing losses that have mostly been derived in the context of diffusion-based generative modeling. This approach usually relies on the concept of time-reversal of SDEs.
To be more precise, the idea is to consider the two controlled SDEs
\begin{align}
\label{eq: def X^u}
    \mathrm d X_s^u &= (f + \sigma u)(X_s^u, s)\,\mathrm{d}s + \sigma(s) \,\mathrm{d}W_s, && X^u_0 \sim p_\mathrm{prior},  \\
\label{eq: def Y^v}
    \mathrm d Y_s^v &= (-\cev{f} + \cev{\sigma} \cev{v})(Y_s^v, s)\,\mathrm{d}s + \cev{\sigma}(s) \,\mathrm{d}W_s,  && Y^v_0 \sim p_\mathrm{target}, 
\end{align}
where $f \in C(\R^d \times [0, T], \R^d)$ and $\sigma \in C(\R^d \times [0, T], \R^{d \times d})$ are fixed and the control functions $u$ and $v$ are learned such that at the optimum $u=u^*$ and $v=v^*$, $X^{u^*}$ is the time-reversal of $Y^{v^*}$, see \cite{richter2023improved}. Clearly, if the time-reversal property is fulfilled, we have $X_T^{u^*} \sim p_\mathrm{target}$ and thus solved our sampling problem. The above setting corresponds to a general bridge between the prior and target density and -- just like in our general setting in \Cref{sec: general evolution} -- has infinitely many solutions. Two ways to attain uniqueness are to either set $v = 0$ and choose $f$ suitably such that $p_Y(\cdot, T) \approx p_\mathrm{prior}$, which corresponds to score-based generative modeling, see \cite[Section 3.2]{richter2023improved}, or to constrain to an annealing strategy between $p_\mathrm{prior}$ and $p_\mathrm{target}$, i.e. to prescribe $p_{X^{u^*}}$, see \cite{vargas2023transport}.

Finally, before proving the loss equivalences from \Cref{prop: BSDE loss equivalences}, let us briefly introduce BSDE-based losses. For more details, we refer to \cite{nusken2021interpolating}. BSDE-based losses build on a stochastic representation of the PDE, which is essentially coming from It\^{o}'s formula, which states
\begin{align}
\begin{split}
\label{eq: Ito formula}
        V(X_T, T) - V(X_0, 0) =& \int_0^T \left(\partial_s V + \frac{1}{2}\tr\left(\sigma\sigma^\top \nabla^2 V\right) + \mu \cdot V \right)(X_s, s) \,\mathrm ds \\
        &\qquad\qquad\qquad\qquad\qquad\qquad+ \int_0^T \sigma^\top \nabla V(X_s, s) \cdot  \mathrm dW_s,
\end{split}
\end{align}
where $X$ is defined by the SDE
\begin{equation}
\mathrm d X_s = \mu(X_s,s)  \mathrm ds + \sigma(s) \, \mathrm dW_s.
\end{equation}
Now, for a PDE
\begin{equation}
    \partial_t V + \frac{1}{2}\tr\left(\sigma\sigma^\top \nabla^2 V\right) + \mu\cdot V + h(\cdot, \cdot, V, \nabla V, \nabla^2 V) = 0,
\end{equation}
where $h\in C(\R^d \times [0, T] \times \R \times \R^d \times \R^{d \times d},\R)$ is a possibly nonlinear function that may depend on the solution $V$ and their derivatives, we may turn \eqref{eq: Ito formula} into
\begin{equation}
\label{eq: Ito formula nonlinear function}
        V(X_T, T) - V(X_0, 0) = -\int_0^T h\left(\cdot, \cdot, V, \nabla V, \nabla^2 V) \right)(X_s, s) \,\mathrm ds + \int_0^T \sigma^\top \nabla V(X_s, s) \cdot  \mathrm dW_s.
\end{equation}
The general idea of BSDE-based losses is now to learn an approximation $\widetilde{V} \approx V$ s.t. \eqref{eq: Ito formula nonlinear function} is fulfilled, e.g. via the loss
\begin{align}
\label{eq: general BSDE loss}
\begin{split}
    \mathcal{L}^\mathrm{BSDE}(\widetilde{V}) = \E\Bigg[&\Bigg( \widetilde{V}(X_T, T) - \widetilde{V}(X_0, 0)  + \int_0^T h\left(\cdot, \cdot, \widetilde{V}, \nabla \widetilde{V}, \nabla^2 \widetilde{V}) \right)(X_s, s) \,\mathrm ds \\
    &\qquad\qquad\qquad\qquad\qquad\qquad\qquad\qquad- \int_0^T \sigma^\top \nabla \widetilde{V}(X_s, s) \cdot  \mathrm dW_s \Bigg)^2 \Bigg],
\end{split}
\end{align}
where typically at least one of the values $\widetilde{V}(X_0, 0)$ and $\widetilde{V}(X_T, T)$ can be replaced by the respective boundary values of the PDE. We can now prove \Cref{prop: BSDE loss equivalences}.

\begin{proof}[Proof of \Cref{prop: BSDE loss equivalences}]
\textit{(i)} Let us start with $\mathcal{L}_{\mathrm{logFP}}$ from \Cref{sec: general evolution} and recall the corresponding PDE \eqref{eq: log-FK}, namely
\begin{equation}
\label{eq: log-FK app}
\partial_t V  + \div(\mu) + \nabla V \cdot \mu - \tfrac{1}{2}\|\sigma^\top \nabla V\|^2 - \tfrac{1}{2}\tr(\sigma \sigma^\top \nabla^2 V) = 0.
\end{equation}
Picking $\mu = f + \sigma u^*$, as in the SDE \eqref{eq: def X^u}, we can write
\begin{equation}
\label{eq: log-FK PDE with u and V}
\partial_t V  + \div(f + \sigma u^*) + \nabla V \cdot (f + \sigma u^*) - \tfrac{1}{2}\|\sigma^\top \nabla V\|^2 + \tfrac{1}{2}\tr(\sigma \sigma^\top \nabla^2 V) - \tr(\sigma \sigma^\top \nabla^2 V) = 0.
\end{equation}
Applying the BSDE loss brings
\begin{align}
\begin{split}
\label{eq: BSDE loss general bridge}
    \mathcal{L}_{\mathrm{logFP}}^{\mathrm{BSDE}}&(u, \widetilde{V}) = \\
    \E\Bigg[&\Bigg( \int_0^T \left(\div(f + \sigma u) + \sigma^\top\nabla \widetilde{V} \cdot (u - w) - \tfrac{1}{2}\|\sigma^\top\nabla \widetilde{V} \|^2 - \tr(\sigma \sigma^\top \nabla^2 \widetilde{V}) \right)(X_s^w, s)\mathrm ds  \\
    & \qquad\qquad\qquad\qquad\qquad- \int_0^T \sigma^\top \nabla \widetilde{V} (X_s^w, s) \cdot \mathrm dW_s + \log \frac{p_\mathrm{target}(X_T^w)}{p_\mathrm{prior}(X_0^w)} \Bigg)^2 \Bigg],
\end{split}
\end{align}
where $X^w$ is defined by
\begin{equation}
\mathrm d X_s^w = (f+\sigma w)(X^w_s,s)  \mathrm ds + \sigma(s) \, \mathrm dW_s, \qquad X_0^w \sim p_\mathrm{prior},
\end{equation}
noting that $u$ has been replaced by a generic forward control $w$, see, e.g. \cite[Section 5.2.1]{nusken2021interpolating}. Since the PDE \eqref{eq: log-FK PDE with u and V} depends on the two functions $u^*$ and $V$, the BSDE loss now also depends on two unknowns instead of only one, as defined in \eqref{eq: general BSDE loss}. Now, considering the time-reversed SDE $Y^v$ given by 
\begin{equation}
\mathrm d Y_s^v = (-\cev{f}+\cev{\sigma} \cev{v})(X^w_s,s)  \mathrm ds + \cev{\sigma}(s) \, \mathrm dW_s, \qquad Y^v_0 \sim p_\mathrm{target},
\end{equation}
we recall Nelson's relation
\begin{equation}
\label{eq: Nelson}
    u^* + v^* = \sigma^\top \nabla \log p_{X^{u^*}} = \sigma^\top \nabla V,
\end{equation}
which relates the optimal controls to the solution $V = \log p_{X^{u^*}}$ \cite{nelson1967dynamical}. Inserting \eqref{eq: Nelson} into \eqref{eq: BSDE loss general bridge}, we get
\begin{align}
\begin{split}
\mathcal{L}_{\mathrm{Bridge}}^{\mathrm{BSDE}}(u, v) = \E\Bigg[&\Bigg( \int_0^T \left(\div(f - \sigma v) -(u+v)\cdot\left(w + \frac{v-u}{2}\right) \right)(X_s^w, s)\mathrm ds  \\
    & \qquad\qquad\qquad- \int_0^T (u+v) (X_s^w, s) \cdot \mathrm dW_s + \log \frac{p_\mathrm{target}(X_T^w)}{p_\mathrm{prior}(X_0^w)} \Bigg)^2 \Bigg],
\end{split}
\end{align}
which is the loss derived in \cite{richter2023improved} when replacing the variance with the second moment, see also the comments in \cite[Appendix A.2]{richter2023improved} and \cite{nusken2021solving}.

\textit{(ii)} The equivalence of the BSDE version of the annealed loss, $\mathcal{L}_{\mathrm{logFP}}^\mathrm{anneal,BSDE}$, with  $\mathcal{L}_{\mathrm{CMCD}}^\mathrm{BSDE}$ can be seen by first noting that the PDE \eqref{eq: log-FK app} with fixed $V$ leads to the BSDE loss
\begin{align}
\begin{split}
\label{eq: BSDE loss annealing}
    \!\!\!\!\mathcal{L}_{\mathrm{logFP}}^{\mathrm{BSDE}}(\widetilde{\mu}) = \E\Bigg[&\Bigg( \int_0^T \left(\div(\widetilde{\mu}) + \nabla V \cdot (\widetilde{\mu} - \gamma) - \tfrac{1}{2}\|\sigma^\top\nabla V \|^2 - \tr(\sigma \sigma^\top \nabla^2 V) \right)(X_s^\gamma, s)\mathrm ds  \\
    & \qquad\qquad\qquad\qquad- \int_0^T \sigma^\top \nabla V (X_s^\gamma, s) \cdot \mathrm dW_s + \log \frac{p_\mathrm{target}(X_T^\gamma)}{p_\mathrm{prior}(X_0^\gamma)} \Bigg)^2 \Bigg],
\end{split}
\end{align}
where $X^\gamma$ is defined by
\begin{equation}
\mathrm d X_s^\gamma = \gamma(X^\gamma_s,s)  \mathrm ds + \sigma(s) \, \mathrm dW_s, \qquad X_0^\gamma \sim p_\mathrm{prior},
\end{equation}
noting that $\mu$ has been replaced by a generic forward drift $\gamma$. Adopting to the choices in \cite{vargas2023transport}, we choose $\widetilde{\mu} = \tfrac{1}{2} \sigma \sigma^\top \nabla V + \nabla \phi$, where $\phi \in C(\R^d \times [0,T], \R)$, and note the identity
\begin{equation}
    \tfrac{1}{2}\int_0^T \div(\sigma\sigma^\top \nabla V)(X^\gamma_s,s) \mathrm ds = \int_0^T\sigma^\top \nabla V(X_s^\gamma,s)\circ \mathrm dW_s  -\int_0^T \sigma^\top \nabla V(X_s^\gamma,s)\cdot \mathrm dW_s,
\end{equation}
where the first stochastic integral is the Stratonovich integral. Plugging those choices into \eqref{eq: BSDE loss annealing}, we readily recover the loss in \cite{vargas2023transport} when replacing the variance with the second moment (and taking $\gamma = \widetilde{\mu}$, in which case, however, one must assure that no gradients w.r.t. the drift in the SDE are taken), namely
\begin{align}
\begin{split}
    \mathcal{L}_{\mathrm{CMCD}}^{\mathrm{BSDE}}(\phi) = \E\Bigg[&\Bigg( \int_0^T \left(\Delta \phi -\tfrac{1}{2} \|\sigma^\top \nabla V \|^2 +\nabla V \cdot (\widetilde{\mu} - \gamma)\right)(X_s^\gamma, s)\mathrm ds  \\
    & \qquad\qquad\qquad\qquad - \int_0^T \sigma^\top \nabla V (X_s^\gamma, s) \circ \mathrm dW_s+ \log \frac{p_\mathrm{target}(X_T^\gamma)}{p_\mathrm{prior}(X_0^\gamma)} \Bigg)^2 \Bigg],
\end{split}
\end{align}
see also the comments above. By slightly abusing notation, we may again set $\nabla \phi = \widetilde{\mu}$ such that we can write $\mathcal{L}_{\mathrm{CMCD}}^\mathrm{BSDE}(\widetilde{\mu})$ instead of $\mathcal{L}_{\mathrm{CMCD}}^\mathrm{BSDE}(\phi)$.

\textit{(iii)} Recalling PDE \eqref{eq: FK time-reversal},
\begin{equation}
    \partial_t V  - \div(f) - \nabla V \cdot f + \tfrac{1}{2}\|\sigma^\top \nabla V\|^2 + \tfrac{1}{2}\tr(\sigma \sigma^\top \nabla^2 V) = 0,
\end{equation}
we get the BSDE-based loss
\begin{align}
\begin{split}
\mathcal{L}_{\mathrm{score}}^{\mathrm{BSDE}}(\widetilde{V}) = \E\Bigg[&\Bigg( \int_0^T \left(\div(-f) +\tfrac{1}{2}\|\sigma^\top \nabla V \|^2 - \sigma^\top \nabla V \cdot w \right)(X_s^w, s)\mathrm ds  \\
    & \qquad\qquad\qquad- \int_0^T \sigma^\top \nabla V (X_s^w, s) \cdot \mathrm dW_s + \log \frac{p_\mathrm{target}(X_T^w)}{p_\mathrm{prior}(X_0^w)} \Bigg)^2 \Bigg],
\end{split}
\end{align}
where $X^w$ is defined by
\begin{equation}
\mathrm d X_s^w = (-f+\sigma w)(X^w_s,s)  \mathrm ds + \sigma(s) \, \mathrm dW_s, \qquad X_0^w \sim p_\mathrm{prior}.
\end{equation}
Making the choice $u = \sigma^\top \nabla V$, the loss then turns into

\begin{align}
\begin{split}
\mathcal{L}_{\mathrm{DIS}}^{\mathrm{BSDE}}(u) = \E\Bigg[&\Bigg( \int_0^T \left(\div(-f) +\tfrac{1}{2}\|u \|^2 - u \cdot w \right)(X_s^w, s)\mathrm ds  \\
    & \qquad\qquad\qquad- \int_0^T u (X_s^w, s) \cdot \mathrm dW_s + \log \frac{p_\mathrm{target}(X_T^w)}{p_\mathrm{prior}(X_0^w)} \Bigg)^2 \Bigg],
\end{split}
\end{align}
which corresponds to the DIS method derived in \cite{berner2022optimal} when taking the second moment instead of the variance, see also \cite[Section 3.2]{richter2023improved}.
\end{proof}

\section{Computational aspects}
\label{sec: computational aspects}

\subsection{Implementation}

\paragraph{Neural networks:} We performed a grid-search over different architecture choices. For the networks $\mu$, $\varphi$, and $\Phi$, we experimented with both Fourier-MLPs as in~\cite{zhang2021path} and standard MLPs with residual connections. In settings where we need to compute Laplacians of our network, we additionally considered the OT-Flow architecture~\cite{onken2021ot,ruthotto2020machine}. For the annealing losses, we parametrize $z$ by a small Fourier-MLP. For the other methods, $z$ does not need to depend on $t$, and we just use a single trainable parameter, see also~\Cref{app: PINN loss details}. For the loss $\mathcal{L}_\mathrm{score}$, we additionally experimented with parametrizations of $\widetilde{V}$ that omit the prior density $p_{\mathrm{prior}}$ and normalizing constant $z$ in~\eqref{eq: V parametrization}, see~\Cref{sec: constrained evolution}.
\paragraph{Hyperparameters:} For all methods, we choose $p_{\mathrm{prior}} = \mathcal{N}(0,\mathrm{I})$ and used domains of the form 
\begin{equation}
    \Omega = \left\{(x,t) \in \R^d \times [0,T] \colon  t\, \underline{\Omega}_{\mathrm{target}} + (1-t)\, \underline{\Omega}_{\mathrm{prior}} \le x \le t\, \overline{\Omega}_{\mathrm{target}} + (1-t)\, \overline{\Omega}_{\mathrm{prior}} \right\},
\end{equation}
where the inequalities are to be understood componentwise. We tuned the rectangular domains of the prior and target distributions $\underline{\Omega}_{\mathrm{prior}},\overline{\Omega}_{\mathrm{prior}},\underline{\Omega}_{\mathrm{target}},\overline{\Omega}_{\mathrm{target}} \in \R^d$ for each problem.
Moreover, we set $\sigma$ to a constant value, i.e., $\sigma(t) = \bar{\sigma}\, \mathrm{I}$.
For the diffusion model, we pick a simple VP-SDE
from~\cite{song2020score} with $f(x,t) \coloneqq -\frac{\bar{\sigma}^2}{2} x$ and sufficiently large $\bar{\sigma}$ and $T$ to ensure that $p_Y(\cdot,T) \approx p_\mathrm{prior}$. For the other methods, we choose $T = 1$ and $\bar{\sigma} \in \{0, \sqrt{2}\}$.

\paragraph{Training and inference:} Each experiment is executed on a single GPU. We train with batch-size $4096$ for $200k$ gradient steps (or until convergence) using the Adam optimizer with an exponentially decaying learning rate. We performed a grid-search over the penalty parameter $\alpha$ of the HJB losses in~\eqref{eq:sb_loss} and~\eqref{eq:ot_loss}, the initial learning rate as well as its decay per step.
We use $100k$ samples to evaluate our methods and simulate our SDEs and ODEs using the Euler-Maruyama and Fourth-order Runge-Kutta (with 3/8 rule) scheme, respectively.

\subsection{Log-likelihoods and importance weights}
\label{sec:likelihood}

This section describes ways to compute the log-likelihood and importance weights for samples $X_T$ obtained from the stochastic process $X$.

\paragraph{ODEs:}
In the setting of normalizing flows, we can compute the evolution of the log-density along the trajectories. Using $\frac{\mathrm{d}}{\mathrm{d}t} X_t = \mu(X_t,t)$ as well as \eqref{eq: log-CE}, one can show that
\begin{equation}
\label{eq:log_likelihood}
   \frac{\mathrm{d}}{\mathrm{d}t} V(X_t,t) =  (\nabla V \cdot \mu -\div(\mu) - \nabla V \cdot \mu)(X_t,t) = -\div(\mu)(X_t, t),
\end{equation}
which is often referred to as the \emph{change-of-variables formula}. Recalling that $V=\log p_X$, we can then compute the (unnormalized) \emph{importance weights} 
\begin{equation}
\label{eq:is}
    w^{(k)} \coloneqq \frac{\rho_\mathrm{target}}{p_{X_T}} (X^{(k)}_T)
\end{equation}
of samples $(X_T^{(k)})_{k=1}^K$ by integrating \eqref{eq:log_likelihood}.

\paragraph{SDEs:}
If we have (an approximation to) the score $\nabla V = \nabla\log p_X$ of an SDE $X$, we can transform it into an ODE with the same marginals using
\begin{equation}
\label{eq:sde_to_ode}
    \mu_{\mathrm{ODE}} = \mu_{\mathrm{SDE}} - \frac{1}{2}\sigma \sigma^\top \nabla V.
\end{equation}
The above relation can be verified via the Fokker-Planck equation~\eqref{eq: FP}, and the resulting ODE is often referred to as \emph{probability flow ODE}~\citep{song2020score}. Note that this also allows us to use the change-of-variables formula in~\eqref{eq:log_likelihood} for SDEs. 

The log-likelihoods can be simulated together with the ODE in~\eqref{eq: ODE} and allow us to compute importance weights in the target space $\R^d$.
If the optimal drift of the SDE can be described via a change of path measures, such as for the annealed flows~\citep{vargas2023transport} or diffusion models~\citep{berner2022optimal}, we can also perform importance sampling in path space $C([0,T],\R^d)$, see, e.g., \cite[Appendix A.12]{berner2022optimal} for further details.

\subsection{Metrics}
\label{app: metrics}

We evaluate the performance of our methods on the following metrics.
\paragraph{Normalizing constants:} We could obtain an estimate $\log z(T)$ of the log-normalizing constant $\log Z$ by our parametrization in~\eqref{eq: V parametrization}. However, since we are interested in the sample quality of our models, we use the log-likelihood to compute a lower bound for $\log Z$, see~\Cref{sec:likelihood}. Note that we do not employ importance sampling for estimating the log-normalizing constant.

\paragraph{Standard deviations:} We also analyze the error when approximating coordinate-wise standard deviations of the target distribution $p_{\mathrm{target}}$, i.e., 
\begin{equation}
       \frac{1}{d} \sum_{k=1}^d \sqrt{\mathbbm{V}[X_{T,i}]}, \quad \text{where}   \quad X_T\sim p_\mathrm{target},
\end{equation}
using samples $(X_T^{(k)})_{k=1}^K$ from our model to approximate the variance. 

\paragraph{Effective sample size:} One would like to have the variance of the importance weights small, or, equivalently, maximize the (normalized) \textit{effective sample size}
\begin{equation}
    \operatorname{ESS} \coloneqq \frac{\left(\sum_{k=1}^K w^{(k)}\right)^2}{ n \sum_{k=1}^K (w^{(k)})^2}.
\end{equation}
The computation of the importance weights is outlined in~\Cref{sec:likelihood}.

\section{Experiments}
\label{sec:app_exp}

\begin{figure}
    \centering
    \begin{subfigure}[t]{\textwidth}
    \centering
    \begin{subfigure}[t]{0.33\textwidth}
         \includegraphics[width=1.0\textwidth,clip]{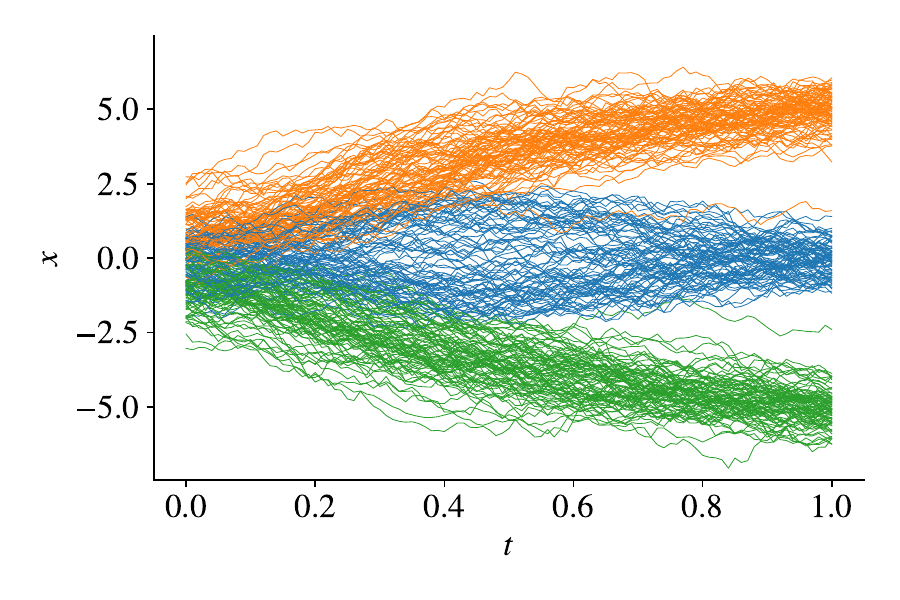}
    \end{subfigure}%
    \begin{subfigure}[t]{0.33\textwidth}
         \includegraphics[width=1.0\textwidth,clip]{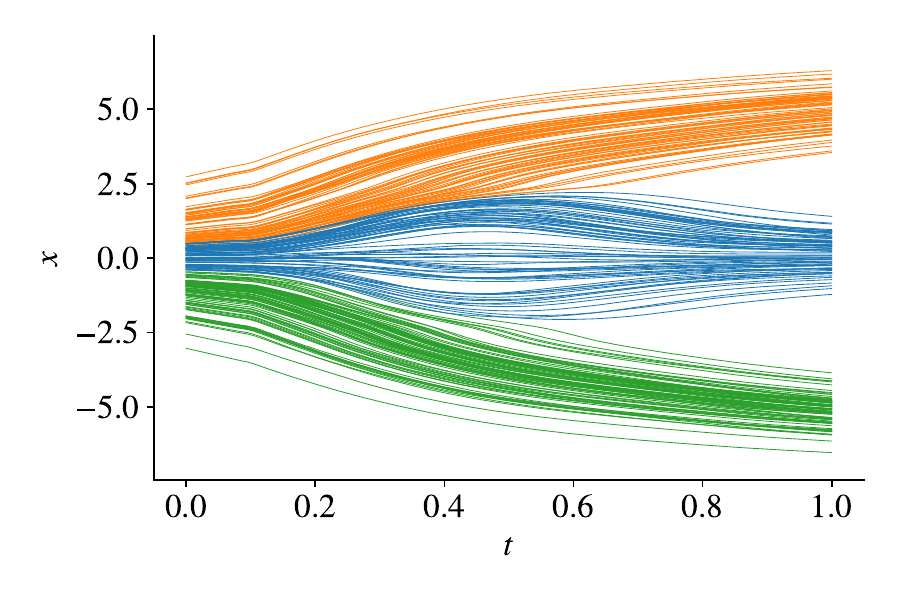}
    \end{subfigure}%
    \begin{subfigure}[t]{0.33\textwidth}
         \includegraphics[width=\textwidth,clip]{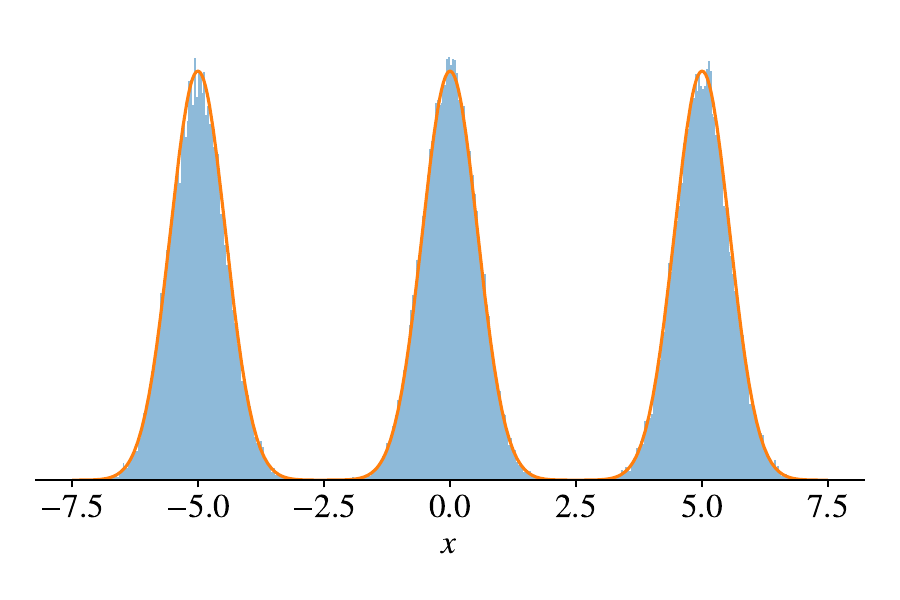}
     \end{subfigure} 
     \vspace*{-0.4em}
     \caption{SDE $(\mathcal{L}_{\mathrm{logFP}})$}
     \vspace*{0.2em}\end{subfigure}\\
     \begin{subfigure}[t]{\textwidth}
     \centering
     \begin{subfigure}[t]{0.33\textwidth}
         \includegraphics[width=1.0\textwidth,clip]{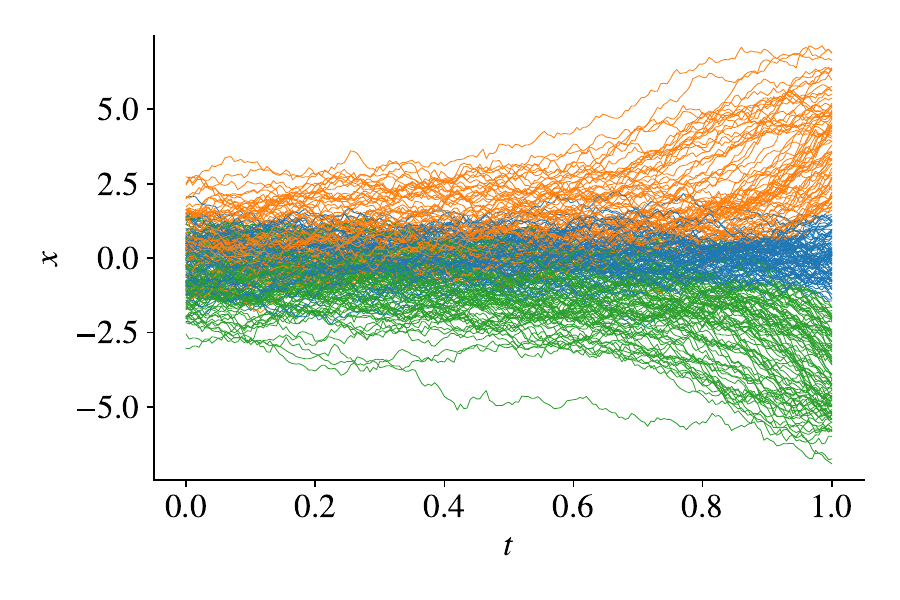}
    \end{subfigure}%
    \begin{subfigure}[t]{0.33\textwidth}
         \includegraphics[width=1.0\textwidth,clip]{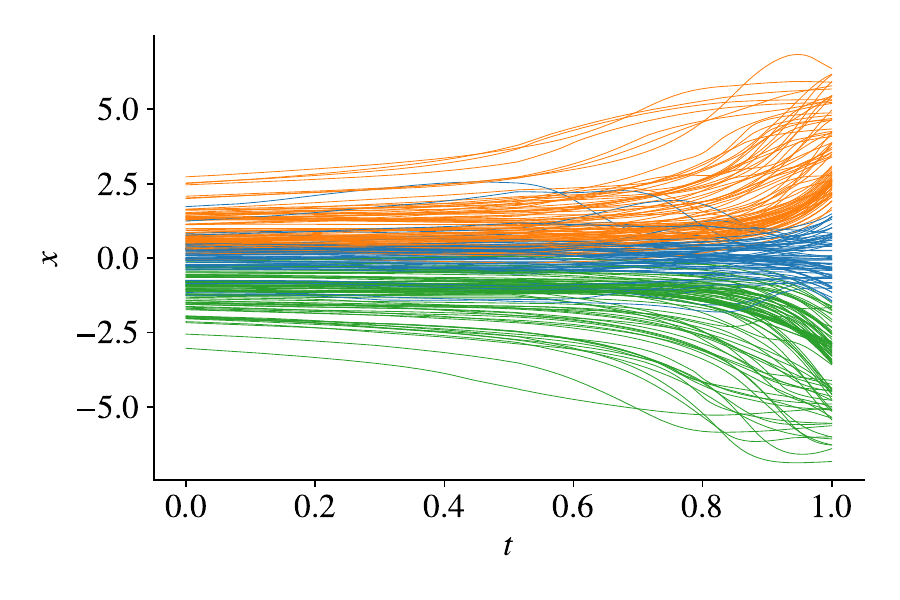}
    \end{subfigure}%
    \begin{subfigure}[t]{0.33\textwidth}
         \includegraphics[width=\textwidth,clip]{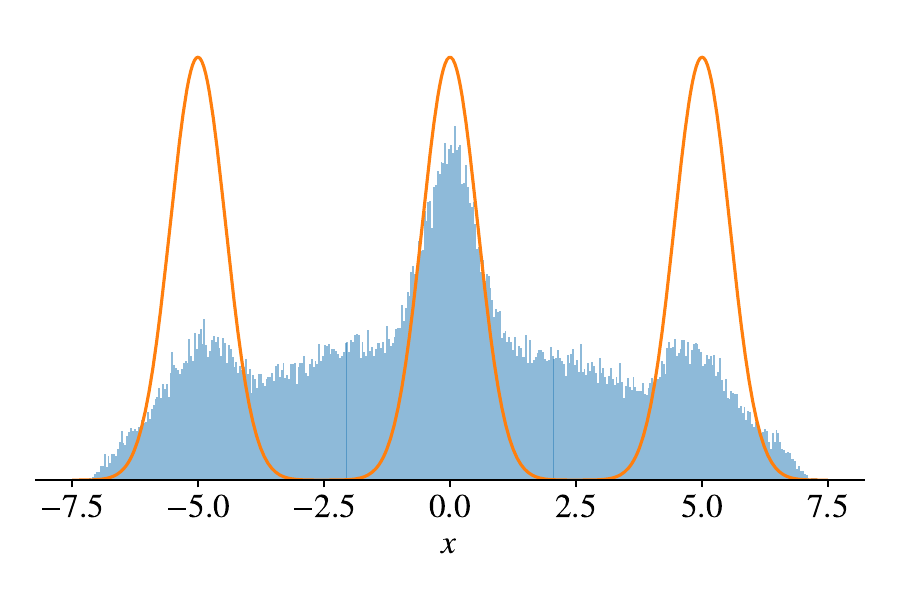}
     \end{subfigure} 
     \vspace*{-0.4em}
     \caption{SDE-anneal $(\mathcal{L}^{\mathrm{anneal}}_{\mathrm{logFP}})$}
     \vspace*{0.2em}
     \end{subfigure}\\
     \begin{subfigure}[t]{\textwidth}
     \centering
     \begin{subfigure}[t]{0.33\textwidth}
         \includegraphics[width=1.0\textwidth,clip]{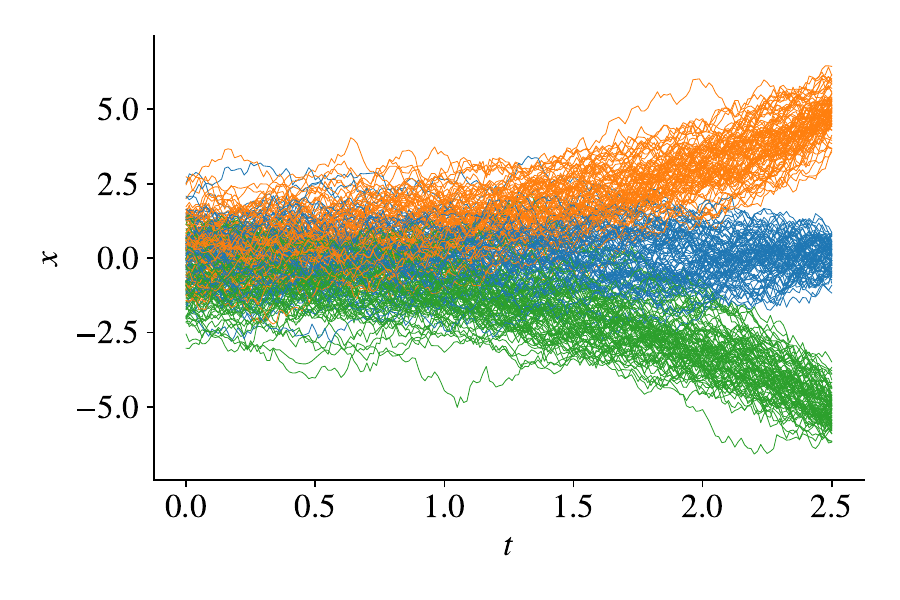}
    \end{subfigure}%
    \begin{subfigure}[t]{0.33\textwidth}
         \includegraphics[width=1.0\textwidth,clip]{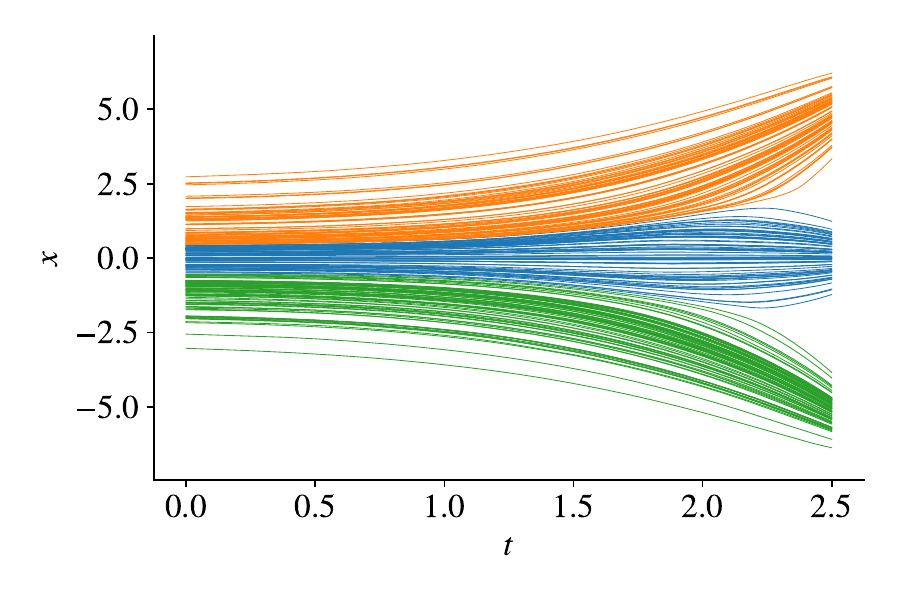}
    \end{subfigure}%
    \begin{subfigure}[t]{0.33\textwidth}
         \includegraphics[width=\textwidth,clip]{{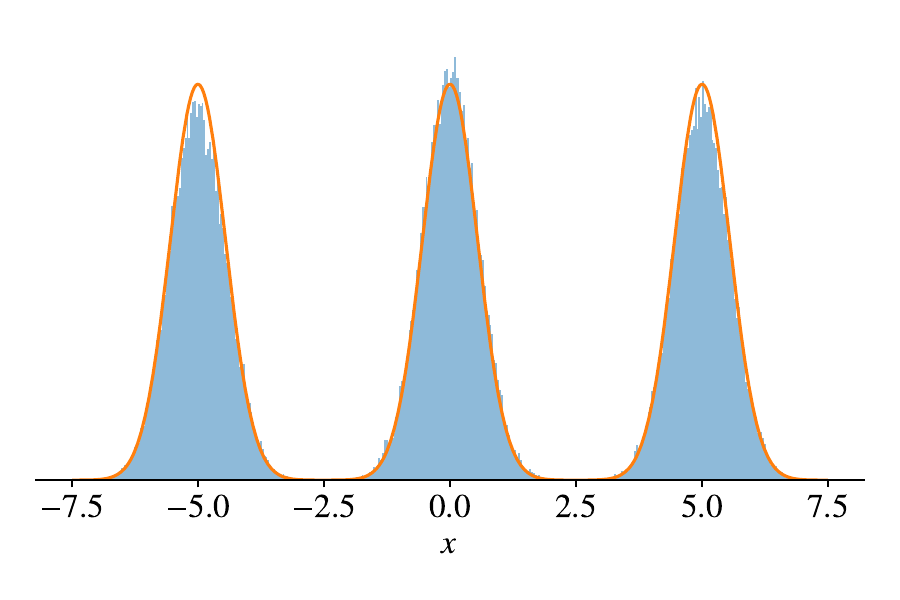}}
     \end{subfigure}%
     \vspace*{-0.4em}
     \caption{SDE-score $(\mathcal{L}_{\mathrm{score}})$}
     \vspace*{0.2em}
     \end{subfigure}\\
     \begin{subfigure}[t]{\textwidth}
     \centering
     \begin{subfigure}[t]{0.33\textwidth}
         \includegraphics[width=1.0\textwidth,clip]{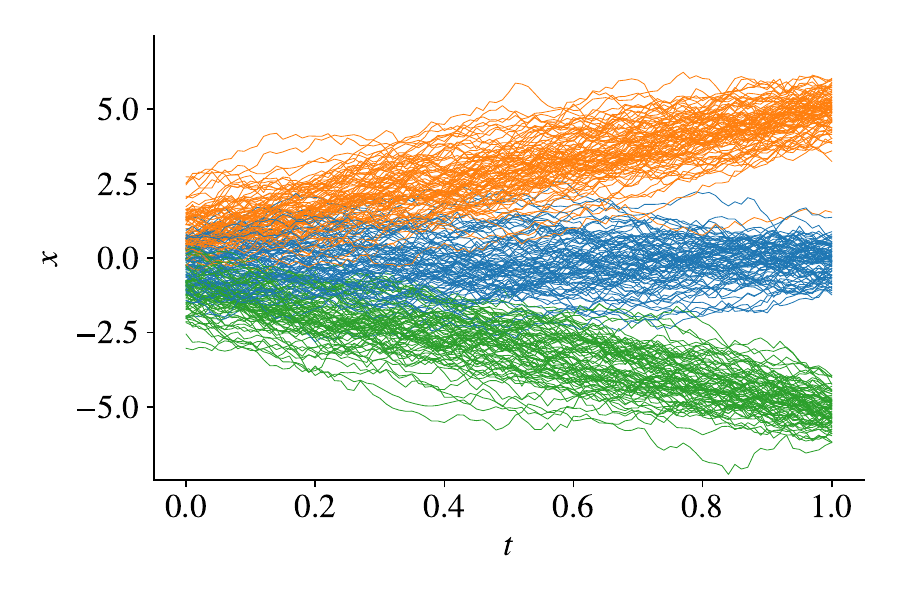}
    \end{subfigure}%
    \begin{subfigure}[t]{0.33\textwidth}
         \includegraphics[width=1.0\textwidth,clip]{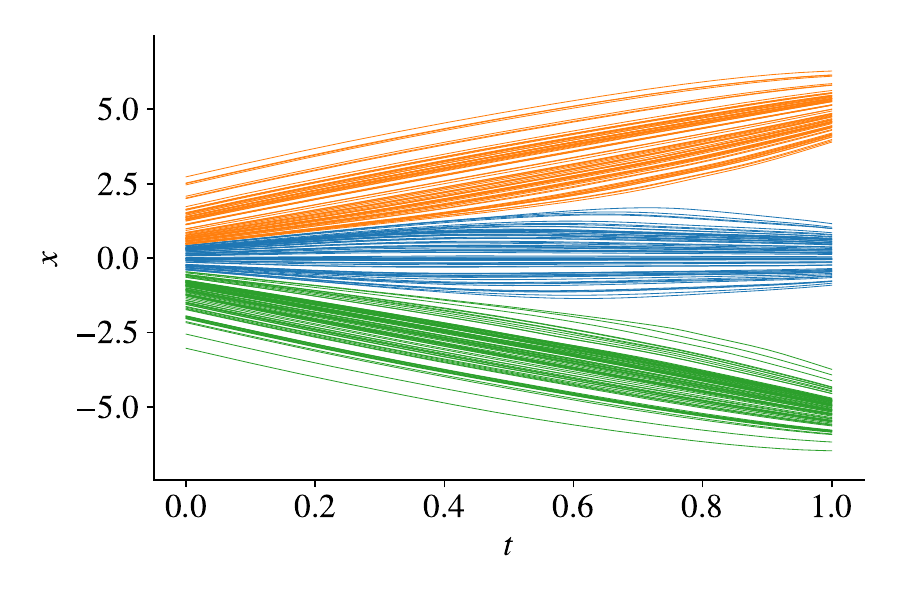}
    \end{subfigure}%
    \begin{subfigure}[t]{0.33\textwidth}
         \includegraphics[width=\textwidth,clip]{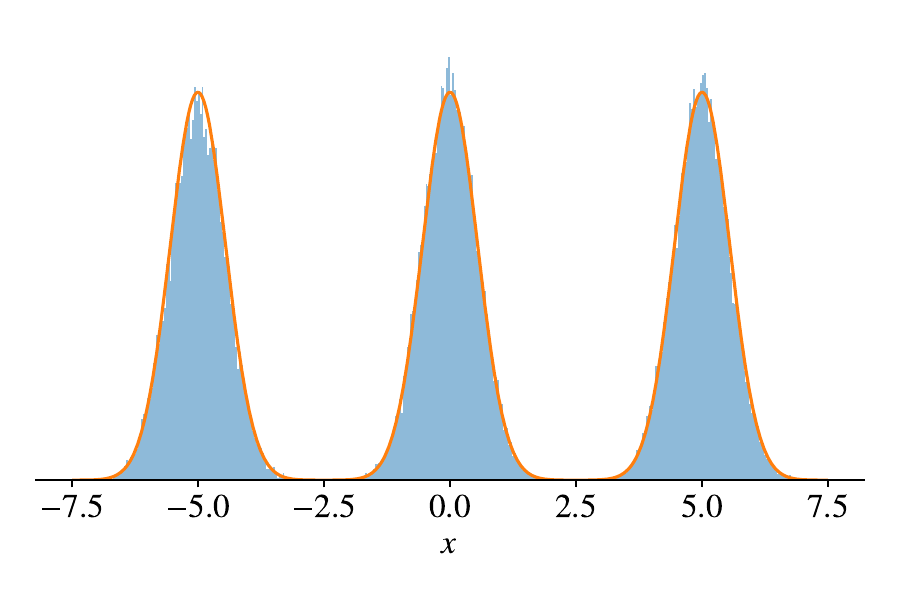}
     \end{subfigure} 
     \vspace*{-0.4em}
     \caption{SB $(\mathcal{L}_{\mathrm{SB}})$}
     \end{subfigure}
     \caption{Trajectories and marginals of our considered SDE-based methods for the GMM example. Note that we also show the corresponding ODE specified in~\eqref{eq:sde_to_ode} that can be used to evaluate the log-likelihoods, see~\Cref{sec:likelihood}. 
     We provide an explanation for the suboptimal performance of $\mathcal{L}^{\mathrm{anneal}}_{\mathrm{logFP}}$ in~\Cref{app: annealing challenges}.}
    \label{fig:marg_sde}
\end{figure}

\begin{figure}
    \centering
     \begin{subfigure}[t]{\textwidth}
     \centering
     \begin{subfigure}[t]{0.33\textwidth}
         \includegraphics[width=1.0\textwidth,clip]{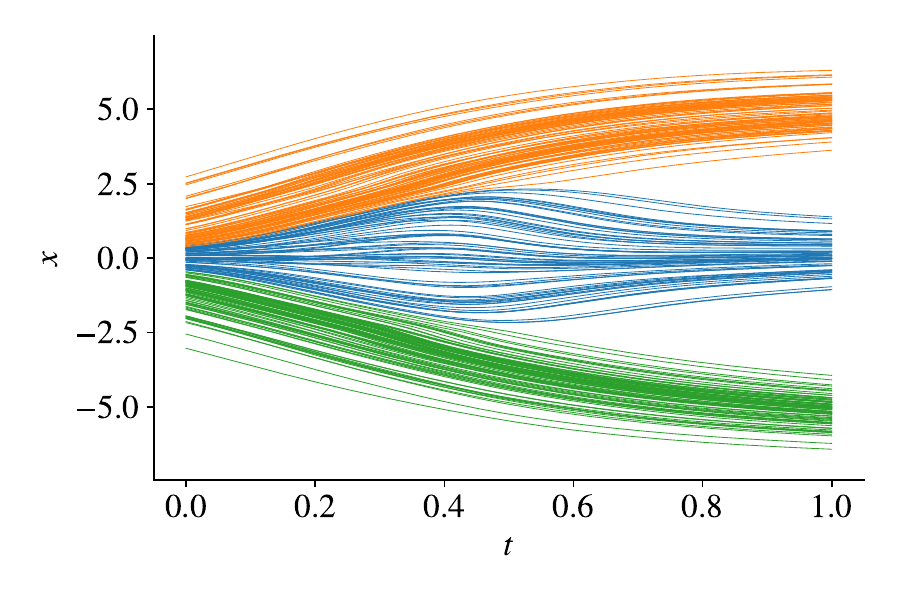}
    \end{subfigure} %
    \hspace{0.1\linewidth}
    \begin{subfigure}[t]{0.33\textwidth}
         \includegraphics[width=\textwidth,clip]{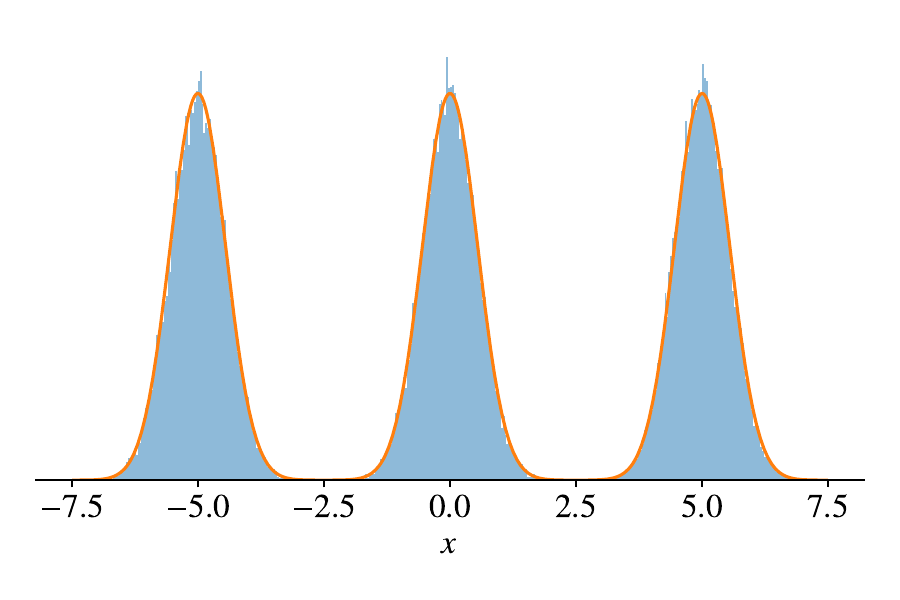}
     \end{subfigure} 
     \vspace*{-0.4em}
     \caption{ODE $(\mathcal{L}_{\mathrm{logCE}})$}
     \vspace*{0.2em}
     \end{subfigure}\\
     \begin{subfigure}[t]{\textwidth}
     \centering
     \begin{subfigure}[t]{0.33\textwidth}
         \includegraphics[width=1.0\textwidth,clip]{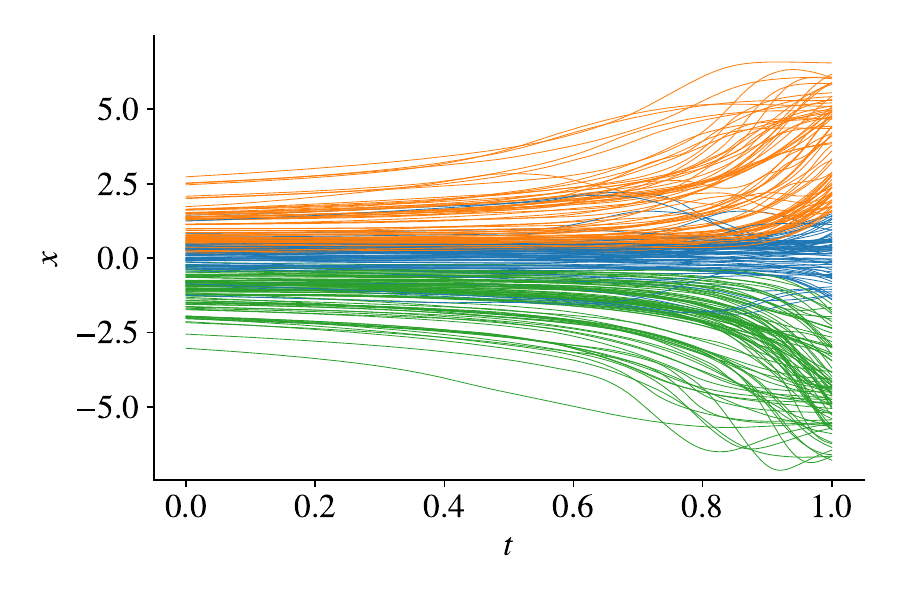}
    \end{subfigure} %
    \hspace{0.1\linewidth}
    \begin{subfigure}[t]{0.33\textwidth}
         \includegraphics[width=\textwidth,clip]{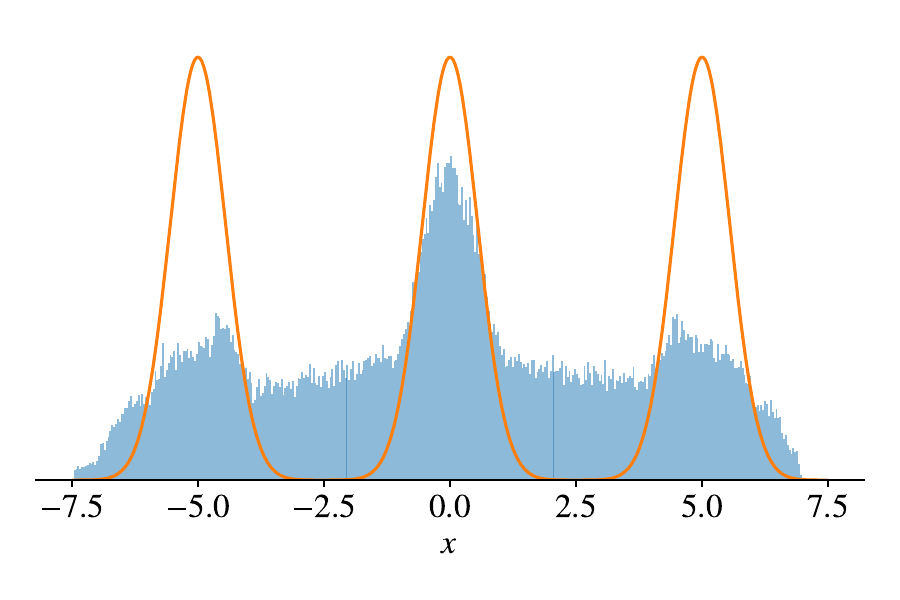}
     \end{subfigure} 
     \vspace*{-0.4em}
     \caption{ODE-anneal $(\mathcal{L}^{\mathrm{anneal}}_{\mathrm{logCE}})$}
     \vspace*{0.2em}
    \begin{subfigure}[t]{0.33\textwidth}
         \includegraphics[width=1.0\textwidth,clip]{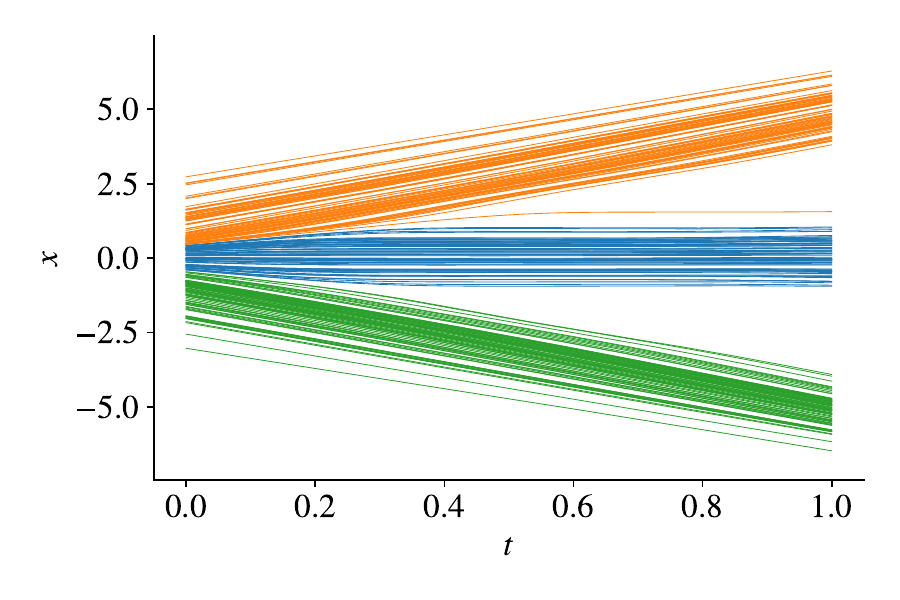}
    \end{subfigure} %
    \hspace{0.1\linewidth}
    \begin{subfigure}[t]{0.33\textwidth}
         \includegraphics[width=\textwidth,clip]{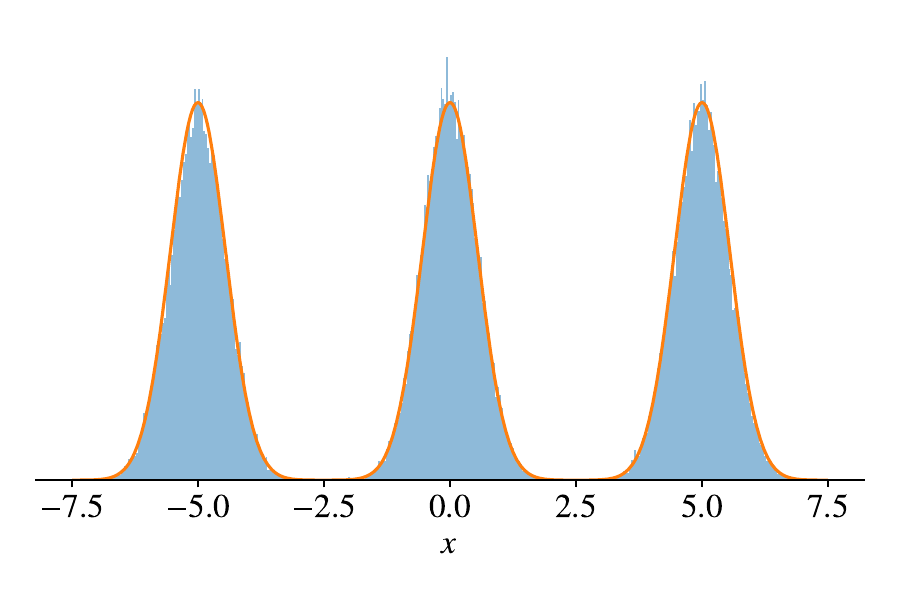}
     \end{subfigure} 
     \vspace*{-0.4em}
     \caption{OT $(\mathcal{L}_{\mathrm{OT}})$}
     \end{subfigure}\\
     \caption{Trajectories and marginals of our considered ODE-based methods for the GMM example. We provide an explanation for the suboptimal performance of $\mathcal{L}^{\mathrm{anneal}}_{\mathrm{logCE}}$ in~\Cref{app: annealing challenges}.}
     \label{fig:marg_ode}
\end{figure}

In the following, we describe our target distributions in more detail.

\paragraph{Gaussian mixture model (GMM):} We consider the density
\begin{equation}
\label{eq: GMM definition}
    \rho_\mathrm{target}(x) =  p_\mathrm{target}(x) = \frac{1}{m} \sum_{i=1}^m  \mathcal{N}(x;\mu_i, \Sigma_i).
\end{equation}
Following~\cite{zhang2021path}, we choose $m=9$, $\Sigma_i=0.3 \,\mathrm{I}$, and \begin{equation}
    (\mu_i)_{i=1}^9= \{-5,0,5\} \times \{-5,0,5\}\subset \R^2
\end{equation}
to obtain well-separated modes. The performance of our considered methods on this target distribution is visualized in~\Cref{fig:marg_sde,fig:marg_ode}.

\textbf{Many-well (MW):}
A typical problem in molecular dynamics considers sampling from the stationary distribution of Langevin dynamics. In our example we shall consider a $d$-dimensional \emph{many-well} potential, corresponding to the (unnormalized) density
\begin{equation}
\label{eq: dw}
   \rho_\mathrm{target}(x) = \exp\left(-\sum_{i=1}^m(x_i^2 - \delta)^2 - \frac{1}{2}\sum_{i=m+1}^{d} x_i^2\right)
\end{equation}
with $m \in \mathbb{N} $ combined double wells and a separation parameter $\delta\in (0, \infty)$, see also~\cite{wu2020stochastic,berner2022optimal}. Note that, due to the many-well structure of the potential, the density contains $2^m$ modes. For these multimodal examples, we can compute reference solutions by numerical integration since $\rho_{\mathrm{target}}$ factorizes in the dimensions.

\subsection{Challenges in annealing strategies}
\label{app: annealing challenges}

As described in \Cref{sec: constrained evolution}, the idea of annealing is to prescribe the solution $p_X$ (or $V := \log p_X$) as a gradual path from $p_\mathrm{prior}$ to $p_\mathrm{target}$. It is not surprising that the actual choice of this path often has a significant effect on the numerical performance of the annealing. In this paper, we use the popular geometric path between the prior and the target, which in log-space can be written as
\begin{equation}
\label{eq: geometric annealing}
   V(\cdot,t) = \tfrac{t}{T} \log \tfrac{\rho_{\mathrm{target}}}{z(t)}  + \left( 1- \tfrac{t}{T}\right) \log p_{\mathrm{prior}},
\end{equation}
cf. \eqref{eq: V parametrization} and noting that $z(t)$ takes care of $p_X$ being a density for each $t \in [0, T]$.

\begin{figure}[!p]
\centering
\begin{subfigure}[b]{\textwidth}
    \centering
   \includegraphics[width=1\linewidth]{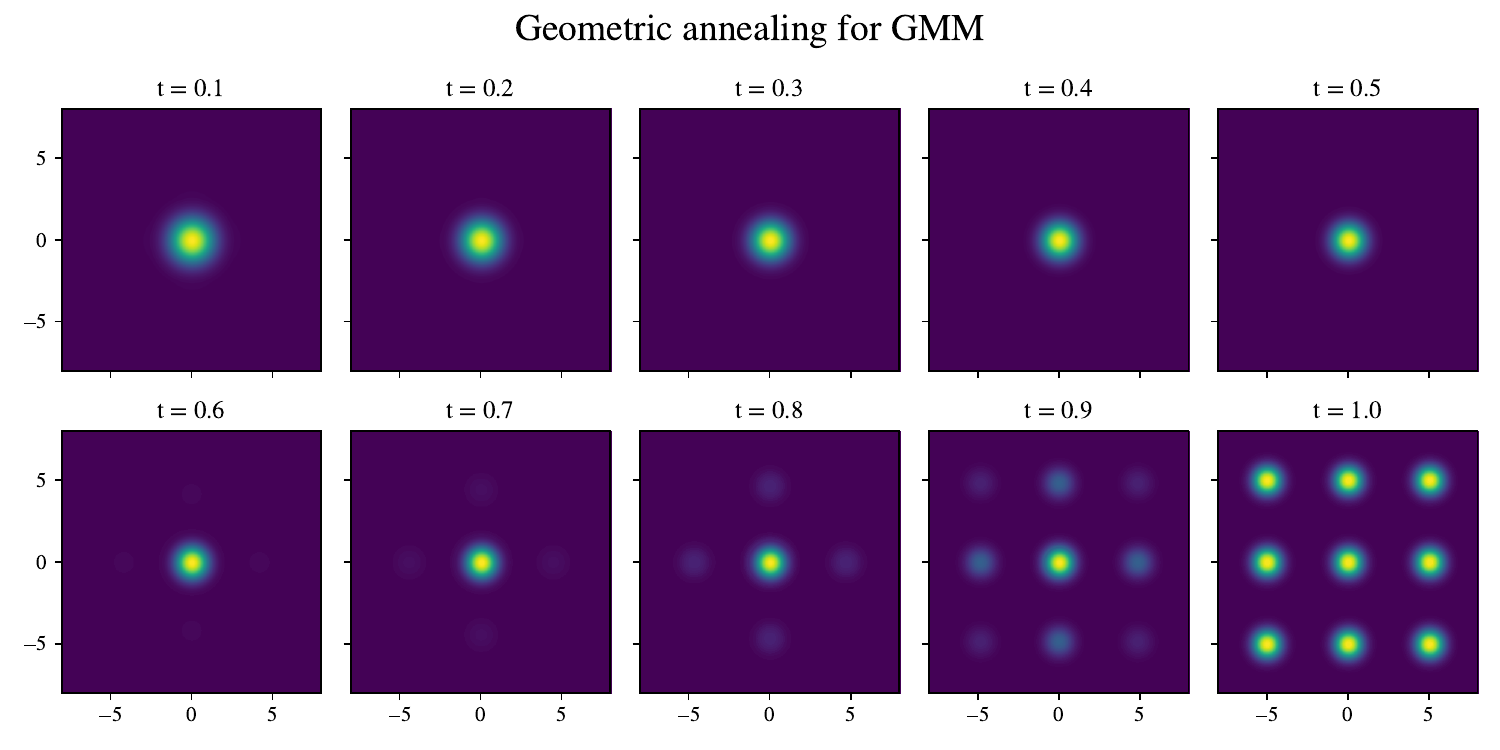}
   \caption{We can observe that the prescribed evolution by the geometric annealing \eqref{eq: geometric annealing} seems to be suboptimal in the sense that most modes of the target only appear late in the annealing path, which might make finding the corresponding drift $\mu$ harder.}
   \label{fig: geometric annealing GMM} 
\end{subfigure} \\
\vspace*{1em}
\begin{subfigure}[t]{\textwidth}
   \centering
   \includegraphics[width=1\linewidth]{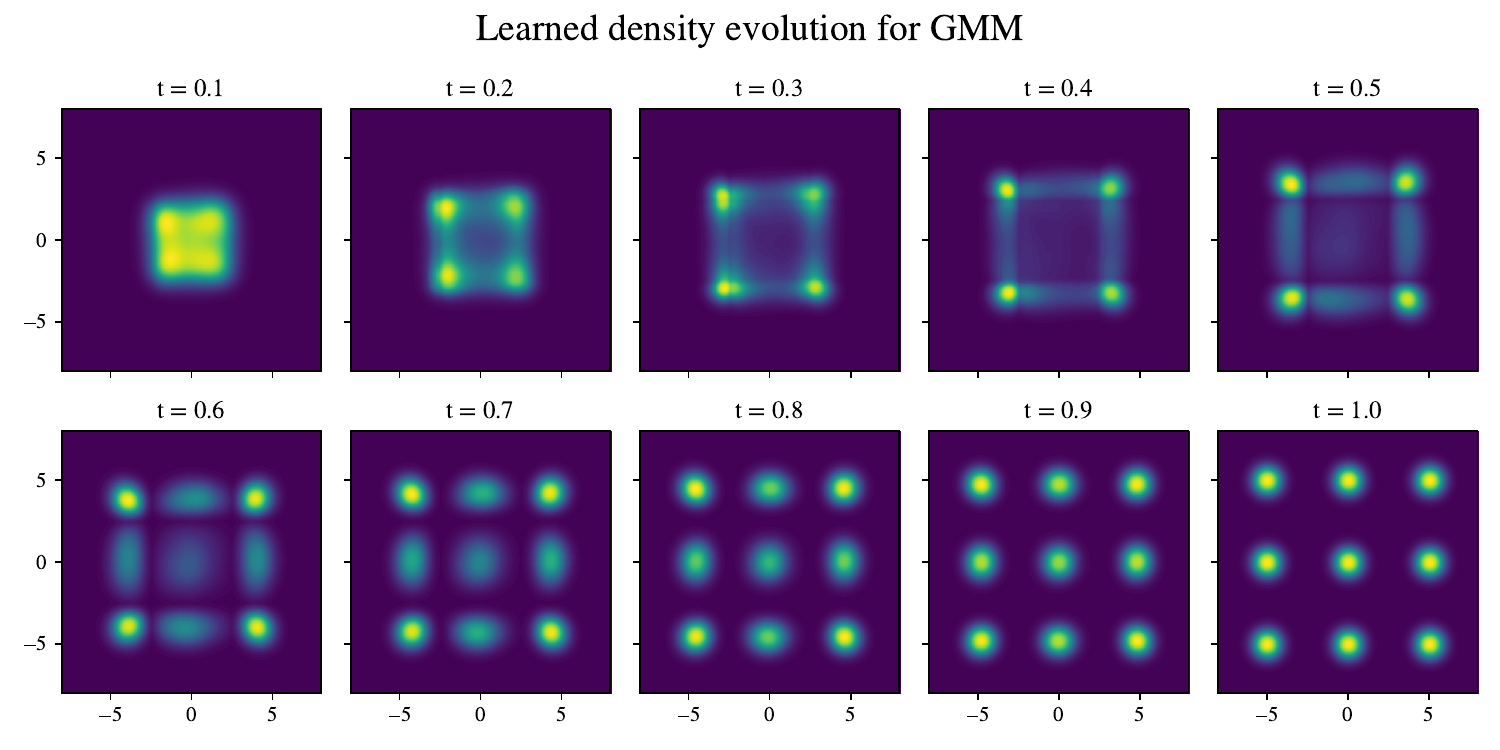}
   \caption{The general loss $\mathcal{L}_\mathrm{logCE}$, on the other hand, optimizes $\mu$ and $V$ simultaneously and thus lets the algorithm find an annealing by itself.}
   \label{fig: learned density evolution GMM}
\end{subfigure} \\
\vspace{0.25em}
\caption{We display different evolutions of the Gaussian prior to the 2-dimensional GMM target defined in \eqref{eq: GMM definition}, once with a prescribed geometric annealing defined in \eqref{eq: geometric annealing} and once learned via the general loss $\mathcal{L}_\mathrm{logCE}$ defined in \eqref{eq: def L_logCE}.}
\label{fig:annealing}
\end{figure}

We have seen in our numerical experiments in \Cref{sec: numerics} and in particular in \Cref{tab:results} that the geometric annealing strategy can lead to more or less satisfying performances, depending on the problem at hand. For the GMM experiment, for instance, the annealing loss performance is rather rather bad, both for the SDE and the ODE. The reason for this can be seen by looking at the density path that is prescribed with \eqref{eq: geometric annealing}, displayed in \Cref{fig: geometric annealing GMM}. We can see that the modes of the target appear only very late in the path, making the task of finding the drift $\mu$ that achieves these densities rather hard. Looking at the path that the non-unique loss $\mathcal{L}_\mathrm{logCE}$ has identified, on the other hand, we realize that the target modes appear much earlier, thus allowing for the identification of the corresponding $\mu$, see \Cref{fig: learned density evolution GMM}. We leave it to further research to come up with more advanced annealing strategies that suffer less from the artifacts described above.

\section{Extensions}

In this section, we mention potential extensions of our framework.

\subsection{Sampling}
\label{sec:sampling}
Let us investigate two choices of how to choose the random variables $(\xi,\tau)$ to penalize the loss in~\eqref{eq: PINN loss short}. We will show how these choices allow us to balance exploration and exploitation. 

\paragraph{Uniform}
We can simply chose $(\xi, \tau) \sim \operatorname{Unif}(\Omega)$ for a sufficiently large compact set $\Omega\subset \R^d \times [0,T]$. This choice allows us to uniformly explore the domain $\Omega$, which is particularly interesting at the beginning of the training. Moreover, different from most other methods, we do not need to rely on (iterative) simulations of the SDE in~\eqref{eq: SDE}.
In order to specify $\Omega$, however, we need prior information to estimate the domain where $V$ is above some minimal threshold. 

\begin{table*}[!t]
\renewcommand{\arraystretch}{1.2}
\setlength{\aboverulesep}{0pt}
\setlength{\belowrulesep}{0pt}
\centering
\caption{Effect of adding additional samples $(\xi,\tau)$ along the trajectories of $X$ for the loss $\mathcal{L}_\mathrm{logCE}$, see~\Cref{sec:sampling}. We simulate and cache $10k$ trajectories of $X$ discretized at $200$ timesteps every $5k$ gradient steps. In every gradient step, we then compute the loss using a random subset of $4096$ samples from the cache and $4096$ uniformly distributed samples. Using only half the number of iterations, i.e., $100k$, we can still improve upon the metrics in~\Cref{tab:results}. The arrows $\uparrow$ and $\downarrow$ indicate whether we want to maximize or minimize a given metric.}
\resizebox{\textwidth}{!}{
\begin{tabular}{lllrrrrr}
    \toprule
     Problem & Sampling & Loss $\downarrow$  & \hspace{-1em} $\Delta\log Z\downarrow$ & \hspace{0.05em} $\mathcal{W}^2_\gamma \downarrow$ & \hspace{-0.4em} $1-\operatorname{ESS} \downarrow$ & \hspace{-0.8em} $\Delta\operatorname{std} \downarrow$ & sec./it. $\downarrow$  \\
    \midrule
    GMM & Uniform & 4.62e-4 & 3.73e-5 & 2.03e-2 & 3.15e-5 & 3.16e-3 & \textbf{0.007} \\
    {\scriptsize $(d=2)$} & Uniform+Traj.    & \textbf{2.05e-4} & \textbf{3.11e-6} &             2.03e-2 &                  \textbf{4.53e-6}         &       \textbf{1.71e-3}     &  0.027 \\ 
    \midrule
    MW & Uniform
    & 3.27e-3 & 8.79e-5 & 1.18e-1 &    6.62e-4 &       3.06e-4 & \textbf{0.008} \\ 
    {\scriptsize $(d=5,m=5,\delta=4)$} & Uniform+Traj.
    & \textbf{3.19e-3} & \textbf{4.40e-5} & 1.18e-1 &    \textbf{2.54e-4} &  3.06e-4 & 0.029 \\ 
    \midrule
    MW
    & Uniform & \textbf{4.83e-2}  &  3.43e-3 & 6.82 & 6.31e-3 & 2.10e-3 & \textbf{0.023} \\
    {\scriptsize $(d=50,m=5,\delta=2)$} 
    & Uniform+Traj. & 3.05e-1  &  \textbf{2.17e-3} &             6.82 &                           \textbf{3.70e-3} &             \textbf{2.99e-4} & 0.051 \\ 
    \bottomrule
\end{tabular}}
\label{tab:results_traj}
\end{table*}

\paragraph{Along the Trajectories}
We can also simulate the SDE using the partially learned drift coefficient $\mu$ to exploit the learned dynamics. This corresponds to the choices $\tau \sim \operatorname{Unif}([0,T])$ and $\xi\sim X_\tau$. Note that we just use the SDE/ODE for sampling the collocation points, and we are not backpropagating through the solver (to update the drift $\mu$). In other words, we detach $\xi$ from the computational graph. In~\Cref{tab:results_traj}, we show that this can lead to faster and better convergence.
Instead of using the drift $\mu$, one could alternatively sample $\xi$ according to $\exp (\widetilde{V}(\cdot, \tau))$, i.e., the current approximation of the density of $X_\tau$, using other sampling methods, such as the Metropolis-adjusted Langevin algorithm. 

Moreover, we want to mention improved sampling strategies for PINNs, see, e.g.,~\cite{tang2023pinns,chen2023adaptive}.
Similar to Quasi-Monte Carlo methods, one could also leverage low-discrepancy samplers for the time coordinate $\tau$, as, e.g., used by~\cite{kingma2021variational}.

\subsection{PINNs} We can make use of a plethora of tricks that have been proposed to stabilize the training of PINNs~\citep{wang2023expert}. For instance, for the networks, one could additionally consider random weight factorization and Fourier features for the spatial coordinates. Moreover, we can choose the penalty parameter $\lambda$ for the HJB loss $\mathcal{L}_{\mathrm{HJB}}$ adaptively based on the residuals and their gradients. Finally, we could also explore the OT-Flow architecture for $\Phi$, which has been successfully employed by~\cite{onken2021ot,koshizuka2022neural,ruthotto2020machine}.

\subsection{Noise Schedule}
\label{sec:noise_schedule}
We can consider time-dependent diffusion coefficients $\sigma$, which have been successfully employed for diffusion models. For instance, we can adapt the VP-SDE in~\cite{song2020score} with    
\begin{equation}
\cev{\sigma}(t) \coloneqq  \sqrt{2\beta(t)} \ \mathrm{I} \quad \text{and} \quad  \cev{f}(x,t) \coloneqq  - \beta(t)x,
\end{equation}
where 
\begin{equation}
     \beta(t) \coloneqq \frac{1}{2}\left(\left(1-\frac{t}{T}\right)\sigma_{\mathrm{min}} + \frac{t}{T} \sigma_{\mathrm{max}}\right).
\end{equation}
Our framework also allows for diffusion coefficients $\sigma$, which depend on the spatial coordinate $x$. Finally, we could also learn the diffusion, for instance, using the parametrization $\sigma=\operatorname{diag}(\exp(s))$ for a neural network $s$.

\subsection{Mean-Field Games}
More generally, we could extend our framework to (stochastic) mean-field games (MFG), mean-field control problems, and generalized SBs using the objective
 \begin{equation}
 \label{eq:objective}
    \mathcal{L}(\mu,\sigma) = \E  \left[\int_{0}^T L\big(X(t),t,\mu(X(t),t)\big) \, \mathrm{d}t \right] + G(p_T),
 \end{equation}
 see~\citep{benamou2017variational,zhang2023mean,liu2022deep,lin2021alternating,koshizuka2022neural,ruthotto2020machine}. In the above, the \emph{Lagrangian} $L$ defines the \emph{running costs}, and the function $G$ specifies the \emph{terminal costs} at time $T$.

\end{document}